\documentclass[a4paper,12pt]{extarticle}
\usepackage{geometry}
\usepackage{amsmath, amsthm, amssymb}

\usepackage{lscape}
\usepackage{fullpage}
\usepackage{csquotes}
\usepackage{cite}
\usepackage{algorithm, algorithmic}
\usepackage{enumitem}
\usepackage{epstopdf}
\usepackage{color}
\usepackage{epsfig}
\usepackage{flexisym}
\usepackage{amsmath}
\usepackage{algorithm, algorithmic}
\usepackage[section]{placeins}
\usepackage[lofdepth,lotdepth]{subfig}
\usepackage{mathtools}

\DeclarePairedDelimiterX\Basics[1](){ #1}
\usepackage{mathtools}
\DeclarePairedDelimiter{\ceil}{\lceil}{\rceil}
\DeclarePairedDelimiter{\floor}{\lfloor}{\rfloor}
\newcommand{\RNum}[1]{\lowercase\expandafter{\romannumeral #1\relax}}

\theoremstyle{plain}

\newtheorem{thm}{Theorem}
\newtheorem{lem}[thm]{Lemma}
\newtheorem{cor}[thm]{Corollary}
\theoremstyle{definition}
\newtheorem{dfn}[thm]{Definition}

\theoremstyle{remark}
\newtheorem{rem}[thm]{Remark}

\newcommand\Tau{\mathbb{T}}
\newcommand\onet{\mathbb{U}}
\newcommand{\m}{\text{max}}
\usepackage{indentfirst}
\usepackage[toc,page]{appendix}

\providecommand{\keywords}[1]{\textbf{\textit{Keywords and phrases}} #1}
\title{
Near-optimal sample complexity for convex tensor completion}
\author{Navid Ghadermarzy, Yaniv Plan, \"{O}zg\"{u}r Y{\i}lmaz\\
Department of Mathematics\\
University of British Columbia}

\begin{document}
\maketitle
  \tableofcontents\newpage
 \begin{abstract}
We analyze low rank tensor completion (TC) using noisy measurements of a subset of the tensor. Assuming a rank-$r$, order-$d$, $N \times N \times \cdots \times N$ tensor where $r=O(1)$, the best sampling complexity that was achieved is $O(N^{\frac{d}{2}})$, which is obtained by solving a tensor nuclear-norm minimization problem. However, this bound is significantly larger than the number of free variables in a low rank tensor which is $O(dN)$. In this paper, we show that by using an atomic-norm whose atoms are rank-$1$ sign tensors, one can obtain a sample complexity of $O(dN)$. Moreover, we generalize the matrix max-norm definition to tensors, which results in a max-quasi-norm (max-qnorm) whose unit ball has small Rademacher complexity. We prove that solving a constrained least squares estimation using either the convex atomic-norm or the nonconvex max-qnorm results in optimal sample complexity for the problem of low-rank tensor completion. Furthermore, we show that these bounds are nearly minimax rate-optimal. We also provide promising numerical results for max-qnorm constrained tensor completion, showing improved recovery results compared to matricization and alternating least squares.
 \end{abstract}
\section{Introduction}\label{introduction}
Representing data as multi-dimensional arrays, i.e., tensors, arises naturally in many modern applications such as interpolating large scale seismic data \cite{kreimer2013tensor,da2015optimization}, medical images \cite{mocks1988topographic}, data mining \cite{acar2005modeling}, image compression \cite{shashua2001linear,liu2013tensor}, hyper-spectral image
analysis \cite{li2010tensor}, and radar signal processing \cite{nion2010tensor}. A more extensive list of such applications can be found in \cite{kolda2009tensor}. There are many reasons where one may want to work with a subset of the tensor entries; (\RNum{1}) often, these data sets are large and we wish to store only a small number of the entries (compression); (\RNum{2}) In some applications, the acquisition of each entry can be expensive, e.g., each entry may be obtained by solving a large PDE \cite{van2013fast}; (\RNum{3}) some of the entries might get lost due physical constraints while gathering them. These restrictions result in situations where one has access only to a subset of the tensor entries. The problem of tensor completion entails recovering a tensor from a subset of its entries.  Without assuming further structure on the underlying tensor, there is no hope of recovering the missing entries as they are independent of the observed entries. Therefore, here (and in many applications) tensors of interest are the ones that can be expressed approximately as a lower dimensional object, compared to the ambient dimension of the tensor. In particular, in this paper we consider tensors that have low CP-rank \cite{carroll1970analysis,harshman1970foundations}. The low rank assumption makes tensor completion a feasible problem. For example,  an order-$d$, rank-$r$ tensor, which has size $N_1 \times N_2 \times \cdots N_d$ where $N_i=O(N)$, has $O(rNd)$ free variables, which is much smaller than $N^d$, the ambient dimension of the tensor.\newline

Tensor completion problem focuses on two important goals: Given a low-rank tensor: (\RNum{1}) identify the sufficient number of entries to be observed in order to recover a good approximation of the tensor, as a function of the size parameters $N_i$, the (CP) rank
$r$, and the order $d$; and (\RNum{2}) design stable and tractable methods that recover the tensor using a subset of its entries.\newline

%

The order-$2$ case, known as \emph{matrix completion}, has been extensively studied in the literature \cite{fazel2002matrix,srebro2005maximum, candes2009exact,keshavan2009matrix,davenport2016overview}. The most basic idea is finding the matrix with lowest rank that is consistent with the measurements. However, rank minimization is NP-hard; therefore, extensive research has been done to find tractable alternatives. The most common approach is using nuclear norm, also known as trace-norm, which is the convex relaxation of the rank function \cite{fazel2002matrix}. It was shown in \cite{candes2010power} that solving a nuclear-norm minimization problem would recover a rank-$r$, $N \times N$ matrix from only $O(rN\text{polylog}(N))$ samples under mild incoherence conditions on the matrix. The nuclear norm is the sum of the singular values of the matrix and it is also the dual of the spectral norm. Extensive research has been done in analyzing variants of nuclear-norm minimization and designing efficient algorithms to solve it as shown in \cite{cai2010singular, candes2010power,candes2010matrix,xu2012alternating}.\newline

An alternative interpretation of the rank and the nuclear-norm of a matrix is based on the minimum number of columns of its factorizations. In particular, the rank of a matrix $M$ is the minimum number of columns of the factors $U,\ V$ where $M=UV'$; the nuclear norm is the minimum product of the Frobenius norms of the factors, i.e., $\|M\|_{\ast} := \text{min} \|U\|_F \|V\|_F\ \text{subject to } M=UV'$ \cite{srebro2005maximum}. An alternative proxy for the rank of a matrix is its max-norm defined as $\|M\|_{\m}:=\text{min} \|U\|_{2,\infty}\|V\|_{2,\infty}$ $\text{subject to}\ M=UV'$ \cite{srebro2005maximum}. The max-norm bounds the norm of the rows of the factors $U$ and $V$ and was used for matrix completion in \cite{foygel2011concentration}. There, the authors studied both max-norm and trace-norm matrix completion by analyzing the Rademacher complexity of the unit balls of these norms. They proved that under uniformly random sampling, either with or without replacement, $m=O(\frac{rN}{\epsilon}\log^3(\frac{1}{\epsilon}))$ samples are sufficient for achieving mean squared recovery error $\epsilon$ using max-norm constrained estimation and $m=O(\frac{rN\log(N)}{\epsilon}\log^3(\frac{1}{\epsilon}))$ samples are sufficient for achieving mean squared recovery error $\epsilon$ using nuclear-norm constrained estimation.\newline



Despite all the powerful tools and algorithms developed for matrix completion, tensor completion problem is still fairly open and not as well understood. For instance, there is a large gap between theoretical guarantees and what is observed in numerical simulations. This is mainly due to the lack of efficient orthogonal decompositions, low-rank approximations, and limited knowledge of structure of low-rank tensors compared to matrices. This large gap has motivated much research connecting the general tensor completion problem to matrix completion by rearranging the tensor as a matrix, including the sum of nuclear-norms (SNN) model that minimizes the sum of the nuclear-norm of matricizations of the tensor along all its dimensions, leading to sufficient recovery with $m=O(r N^{d-1})$ samples \cite{liu2013tensor,gandy2011tensor}. More balanced matricizations, such as the one introduced in \cite{mu2014square}, can result in a better bound of $m=O(rN^{\ceil{\frac{d}{2}}})$ samples.\newline

Once we move from matrices to higher order tensors, many of the well-known facts of matrix algebra cease to be true. For example, even a best rank-$k$ approximation may not exist for some tensors, illustrated in \cite[Section 3.3]{kolda2009tensor}, showing that the space of tensors with rank at most $2$ is not closed. Interestingly there is a paper titled  ``\emph{Most tensor problems are NP-hard}" \cite{hillar2013most} which proves that many common algebraic tasks are NP-hard for tensors with $d \geq 3$, including computing the rank, spectral norm, and nuclear norm. Computational complexity of directly solving tensor completion and the inferior results of matricization make tensor completion challenging.\newline

Having all these complications in mind, on the theoretical side, a low-rank tensor has $O(rdN)$ free variables, but an upper-bound of $O(rN^{\ceil{\frac{d}{2}}})$ on the sample complexity. When $d>2$, the polynomial dependence on $N$ seems to have a lot of room for improvement. Moreover, it is well-known that empirical recovery results are much better when the tensor is not rearranged as a matrix, even though these results are attempting to solve an NP-hard problem. This has resulted in efforts towards narrowing this gap, including heuristic algorithms \cite{liu2013tensor,bazerque2013rank}. In spite of good empirical results and reasonable justifications, a theoretical study filling in the gap was not presented in these cases.\newline

Nuclear norm of a tensor, defined as the dual of the spectral norm was originally formulated in \cite{schatten1985theory,grothendieck1955produits} and has been revisited more in depth in the past few years, e.g., in \cite{derksen2013nuclear,hu2015relations}. Recently \cite{yuan2016tensor} studied tensor completion using nuclear-norm minimization and proved that under mild conditions on the tensor, $m=O(\sqrt{r}N^{\frac{d}{2}} \log(N))$ measurements is sufficient for successful recovery, but this is still far away from the number of free variables.\newline

In effort to obtain linear dependence on $N$, we analyze tensor completion using a max-qnorm (max-quasi-norm) constrained algorithm where the max-qnorm is a direct generalization of the matrix max-norm to the case of tensors. Unfortunately, max-qnorm is non-convex. However, analyzing the unit-ball of the dual of the dual of the max-qnorm (which is a convex norm) led us to define and analyze a convex atomic-norm ( which we call M-norm) constrained least squares problem, where we obtained optimal recovery bounds on the size of the tensor. The main contribution of this paper is as follows. Consider an order-$d$ tensor $T \in \mathbb{R}^{N_1 \times \cdots \times N_d}$ where $N_i = O(N)$ for $1\leq i \leq d$.
\begin{itemize}
\item We define the M-norm and max-qnorm of tensors as a robust proxy for the rank of a tensor. We prove that both M-norm and max-qnorm of a bounded low-rank tensor is upper-bounded by a quantity that just depends on its rank and its infinity norm and is independent of $N$.\medskip
\item We use a generalization of Grothendieck's theorem to connect the max-qnorm of tensors to its nuclear decomposition with unit infinity-norm factors. Using this, we bound the Rademacher complexity of the set of bounded tensors with low max-qnorm. This also establishes a theoretical framework for further investigation of low max-qnorm tensors.\medskip
\item We prove that, with high probability, $m=O(r^{\frac{3d}{2}} d N)$ (or $m=O(R^2 N)$ if M-norm is bounded by $R$) samples are sufficient to estimate a rank-$r$ bounded tensor using a convex least squares algorithm. Moreover, we derive an information-theoretic lower bound that proves $m=O(R^2N)$ measurements is necessary for recovery of tensors with M-norm less than $R$. This proves that our bound is optimal both in its dependence on $N$ and the M-norm bound $R$. It is worth mentioning though, that the bound we prove in this paper is not necessarily optimal in $r$, the rank of the tensor.\medskip
\item Through synthetic numerical examples, we illustrate the advantage of using algorithms designed for max-qnorm constrained tensor completion instead of algorithms, using matricization. These algorithms significantly improve algorithms based on matricization and alternating least squares (ALS). It is worth mentioning that computing the nuclear norm of a general tensor is known to be NP-hard. Although it is not known whether computing the M-norm or max-qnorm of a tensor is NP-hard or not, our numerical results for max-qnorm constrained least squares, using a simple projected quasi-Newton algorithm give promising results.
\end{itemize}
\subsection{Notations and basics on tensors}
We adopt the notation of Kolda and Bader's review on tensor decompositions \cite{kolda2009tensor}. Below, $\lambda$, $\sigma$, and $\alpha$ are used to denote scalars, and $C$ and $c$ are used to denote universal constants. Vectors are denoted by lower case letters, e.g., $u$, and $v$. Both matrices and tensors are represented by upper case letters, usually using $A$ and $M$ for matrices, and $T$ and $X$ for tensors. Tensors are a generalization of matrices to higher order, also called multi-dimensional arrays. For example, a first order tensor is a vector and a second order tensor is a matrix. $X \in \bigotimes_{i=1}^{d}  \mathbb{R}^{N_i}$ is a $d$-th order tensor whose $i$-th size is $N_i$. We also denote $\bigotimes_{i=1}^{d}  \mathbb{R}^{N}$ as $\mathbb{R}^{N^d}$. Elements of a tensor are either specified as $X_{i_1, i_2, \cdots, i_d}$ or $ X(i_1, i_2, \cdots, i_d)$, where $1 \leq i_j \leq N_j$ for $1 \leq j \leq d$. We also use $X_{\omega}$ as a shorthand to refer the index $\omega$ of a tensor, where $\omega=(i_1, i_2, \cdots, i_d)$ is an $n$-tuple determining the index $X(i_1, i_2, \cdots, i_d)$.\newline

Inner products are denoted by $\langle \cdot , \cdot \rangle$. The symbol $\circ$ represents both matrix and vector outer products where $T=U_1 \circ U_2 \circ \cdots \circ U_d$ means $T(i_1,i_2,\cdots,i_d)=\sum_{k} U_1(i_1,k)U_2(i_2,k)$ $\cdots U_d(i_d,k)$, where $k$ ranges over the columns of the factors. In the special case of vectors, $T=u_1 \circ u_2 \circ \cdots \circ u_d$ means $T(i_1,i_2,\cdots,i_d)=u_1(i_1)u_2(i_2)\cdots u_d(i_d)$. Finally $[N]:=\{1, \cdots, N\}$ is the shorthand notation we use for the set of integers from $1$ to $N$.
\subsubsection{Rank of a tensor}
A unit tensor is a tensor $U \in \bigotimes_{j=1}^{d}  \mathbb{R}^{N_j}$ that can be written as
\begin{equation}\label{rank_one}
U=u^{(1)} \circ u^{(2)} \circ \cdots \circ u^{(d)},
\end{equation}
where $u^{(j)} \in \mathbb{R}^{N_j}$ is a unit-norm vector. The vectors $u^{(j)}$ are called the components of $U$. Define $\onet_d$ to be the set of unit tensors of order $d$. A rank-$1$ tensor is a scalar multiple of a unit tensor.\newline

The rank of a tensor $T$, denoted by rank($T$) is defined as the smallest number of rank-$1$ tensors that generate $T$ as their sum, i.e.,
\begin{equation*}
T = \sum_{i=1}^r \lambda_i U_i = \sum_{i=1}^r \lambda_i u_i^{(1)} \circ u_i^{(2)} \circ \cdots \circ u_i^{(d)},
\end{equation*}
where $U_i \in \onet_d$ is a unit tensor. This low-rank decomposition is also known as  CANDECOMP/PARAFAC (CP) decomposition \cite{harshman1970foundations,carroll1970analysis}. In this paper we use CP decompositions; however, we note that there are other decompositions that are used in the literature such as Tucker decomposition \cite{tucker1966some}. For a detailed overview of alternate decompositions, refer to \cite{kolda2009tensor}.
\subsubsection{Tensor norms}
Define $\Tau_d$ to be set of all order-$d$ tensors of size $N_1 \times N_2 \times \cdots \times N_d$. For $X, T \in \Tau_d$, the inner product of $X$ and $T$ is defined as:
\begin{equation*}
\langle X, T\rangle = \sum_{i_1=1}^{N_1} \sum_{i_2=1}^{N_2} \cdots \sum_{i_d=1}^{N_d} X_{i_1, i_2, \cdots, i_d} T_{i_1, i_2, \cdots, i_d}.
\end{equation*}

Consequently the Frobenius norm of a tensor is defined as
\begin{equation}
\|T\|_F^2 := \sum_{i_1=1}^{N_1} \sum_{i_2=1}^{N_2} \cdots \sum_{i_d=1}^{N_d} T_{i_1, i_2, \cdots, i_d}^2 = \langle T,T \rangle.
\end{equation}

Using the definition of unit tensors one can define the spectral norm of tensors as
\begin{equation}
\|T\| := \underset{U \in \onet_d}{\text{max}} \langle T, U\rangle.
\end{equation}

Similarly, nuclear-norm was also generalized for tensors (see \cite{lim2010multiarray,friedland1982variation}, although the original idea dates back to Grothendieck \cite{grothendieck1955produits}) as
\begin{equation}\label{nuclear_norm}
\|T\|_{\ast} := \underset{\|X\| \leq 1}{\text{max}} \langle T, X\rangle.
\end{equation}

Finally we generalize the definition of max-norm to tensors as
\begin{equation}\label{max_norm_tensor_definition}
\|T\|_{\text{max}}:= \underset{T=U^{(1)} \circ U^{(1)} \circ \cdots \circ U^{(d)}}{\text{min}}\lbrace \prod_{j=1}^{d} \|U^{(j)}\|_{2,\infty}\rbrace,
\end{equation}
where, $\|U\|_{2,\infty} = \underset{\|x\|_2=1}{\text{sup}} \|Ux\|_{\infty}$. In Section \ref{max_norm_section}, we prove that for $d>2$ this generalization does not satisfy the triangle inequality and is a quasi-norm (which we call max-qnorm). We analyze the max-qnorm thoroughly in Section \ref{max_norm_section}.

\subsection{Simplified upper bound on tensor completion recovery error}\label{main_result_section}
Without going into details, we briefly state and compute the upper bounds we establish (in Section \ref{section_Max_norm_constrained_LS_estimation}) on the recovery errors associated with M-norm and max-qnorm constrained tensor completion. For ease of comparison, we assume $N_1=N_2=\cdots=N_d=N$. Given a rank-$r$, order-$d$ tensor $T^{\sharp} \in \bigotimes_{i=1}^{d} \mathbb{R}^N$, and a random subset of its entries with indices in $S=\{\omega_1,\omega_2,\cdots,\omega_m\},\ \omega_i \in [N] \times [N] \times \cdots \times [N]$, we observe $m$ noisy entries $\{Y_{\omega_t}\}_{t=1}^{m}$ of $\{T^{\sharp}(\omega_t)\}_{t=1}^{m}$, where each observation is perturbed by iid noise with mean zero and variance $\sigma^2$. To give a simple version of the result we assume that indices in $S$ are drawn independently at random with the same probability for each observation, i.e., we assume uniform sampling to give a simple theorem. We provide the general observation model in Section \ref{section_observation} and a general version of the theorem (which covers both uniform and non-uniform sampling) in Section \ref{TC_maxnorm_section} and prove it in Section \ref{proof_theorem_atomic_TC}. The purpose of tensor completion is to recover $T^{\sharp}$, from $m$ random samples of $T^{\sharp}$ when $m \ll N^d$.
\begin{thm}\label{theorem_simplified}
Consider a rank-$r$, order-$d$ tensor $T^{\sharp} \in \bigotimes_{i=1}^{d} \mathbb{R}^N$ with $\|T^{\sharp}\|_{\infty} \leq \alpha$. Assume that we are given a collection of noisy observations
$$Y_{\omega_t}= T^{\ast}(\omega_t) + \sigma \xi_t\ , \ \ t=1,\cdots,m,$$
where the noise sequence $\xi_t$ are i.i.d. standard normal random variables and each index $\omega_t$ is chosen uniformly random over all the indices of the tensor. Then if $m>dN$, there exist a constant $C<20$ such that the solution of
\begin{equation}
\hat{T}_{M} = \underset{X}{\text{arg min }} \frac{1}{m}\sum_{t=1}^{m} (X_{\omega_t}-Y_{\omega_t})^2 \ \ \ \ \text{subject to}\ \ \ \  \|X\|_{\infty} \leq \alpha,\ \|X\|_{M} \leq (r\sqrt{r})^{d-1} \alpha,
\end{equation}
satisfies
$$\frac{\|T^{\sharp}-\hat{T}_{M}\|_F^2}{N^d} \leq C (\alpha + \sigma) \alpha (r\sqrt{r})^{d-1}\sqrt{\frac{d N}{m}}.$$
with probability greater than $1-e^{\frac{-N}{\text{ln}(N)}}-e^{-dN}$. Moreover, the solution of 
\begin{equation}
\hat{T}_{\m} = \underset{X}{\text{arg min }} \frac{1}{m}\sum_{t=1}^{m} (X_{\omega_t}-Y_{\omega_t})^2 \ \ \ \ \text{subject to}\ \ \ \  \|X\|_{\infty} \leq \alpha,\ \|X\|_{\m} \leq (\sqrt{r^{d^2-d}}) \alpha,
\end{equation}
satisfies
 $$\frac{\|T^{\sharp}-\hat{T}_{\m}\|_F^2}{N^d} \leq C_d (\alpha + \sigma) \alpha \sqrt{r^{d^2-d}}\sqrt{\frac{d N}{m}},$$
with probability greater than $1-e^{\frac{-N}{\text{ln}(N)}}-e^{-dN}$. 
\end{thm}
\begin{rem}
Above, $\|X\|_M$ is the M-norm of tensor $X$ which is an atomic norm whose atoms is rank-$1$ sign tensors defined in Section \ref{section_tensor_maxnorm_Mnorm} \eqref{atomoc_norm_definition} and $\|X\|_{\m}$ is max-qnorm of tensor $X$ which is a generalization of matrix max-norm to tensors defined in Section \ref{section_tensor_maxnorm_Mnorm} \eqref{maxnorm_tensor}.
\end{rem}
\begin{rem}[{\bf{theoretical contributions}}]
The general framework for establishing these upper bounds is already available (the key is to control the Rademacher complexity of the set of interest). The methods to adapt this to the matrix case are available in, e.g., \cite{srebro2004learning,cai2016matrix}. To move to tensor completion, we study the interaction of the max-qnorm, the M-norm, and the rank of a tensor in Section \ref{max_norm_section}. The tools given in Section \ref{max_norm_section} allow us to generalize matrix completion to tensor completion. 
\end{rem}
\subsection{Organization}
In Section \ref{related_works}, we briefly overview recent results on tensor completion and max-norm constrained matrix completion. In Section \ref{max_norm_section}, we introduce the generalized tensor max-qnorm and characterize the max-qnorm unit ball that is crucial in our analysis. This also results in defining a certain convex atomic-norm which gives similar bounds on constrained tensor completion problem. We also prove that both the M-norm and the max-qnorm of a bounded rank-$r$ tensor $T$ can be bounded by a function of $\|T\|_{\infty}$ and $r$, independently of $N$. We have deferred all the proofs to Section \ref{section_proofs}. In Section \ref{TC_maxnorm_section}, we explain the tensor completion problem and state the main results on recovering low-rank bounded tensors. We also compare our results with previous results on tensor completion and max-norm constrained matrix completion. In Section \ref{lower_bound_section}, we state an upper bound on the performance of the M-norm constrained tensor completion which proves optimal dependence on the size. In Section \ref{experiments and algorithms}, we present numerical results on the performance of max-qnorm constrained tensor completion and compare it with applying matrix completion on the matricized version of the tensor and Section \ref{section_proofs} contains all the proofs.
\section{Related work}\label{related_works}
\subsection{Tensor matricization}\label{tensor_matricization_section}
The process of reordering the elements of a tensor into a matrix is called matricization, also known as unfolding or flattening. For a tensor $X \in \bigotimes_{i=1}^{d}  \mathbb{R}^{N_i}$, \emph{mode-}$i$ fibers of the tensor are $\Pi_{j \neq i} N_j$ vectors obtained by fixing all indices of $X$ except for the $i$-th one. The \emph{mode-}$i$ matricization of $X$, denoted by $X_{(i)} \in \mathbb{R}^{N_i \times \Pi_{j \neq i} N_j}$ is obtained by arranging all the \emph{mode-}$i$ fibers of $X$ along columns of the matrix $X_{(i)}$. More precisely, $X_{(i)}(i_i,j)=X(i_1, i_2, \cdots, i_d)$, where
$$j=1+\sum_{k=1,k \neq i}^d(i_k-1)J_k\ \ \ \ \text{with}\ \ \ \ J_k=\Pi_{m=1,m\neq i}^{k-1} N_m.$$
A detailed illustration of these definitions can be found in \cite{kolda2006multilinear,kolda2009tensor}.
A generalization of these unfoldings was proposed by \cite{mu2014square} that rearranges $X_{(1)}$ into a more balanced matrix: For $j \in \{1,\cdots,d\}$, $X_{[j]}$ is obtained by arranging the first $j$ dimensions along the rows and the rest along the columns. In particular, using Matlab notation $X_{[j]}=\text{reshape}(X_{(1)},\Pi_{i=1}^{i=j} N_i,\Pi_{i=j+1}^{i=d} N_i)$. More importantly, for a rank-$r$ tensor $T = \sum_{i=1}^r \lambda_i u_i^{(1)} \circ u_i^{(1)} \circ \cdots \circ u_i^{(d)}$, $T_{[j]} = \sum_{i=1}^r \lambda_i (u_i^{(1)} \otimes u_i^{(2)} \otimes \cdots \otimes u_i^{(j)})  \circ (u_i^{(j+1)} \otimes \cdots \otimes u_i^{(d)})$, which is a rank-$r$ matrix. Here, the symbol $\otimes$, represents the Kronecker product. Similarly, the rank of all matricizations defined above are less than or equal to the rank of the tensor.
\subsection{Past results}\label{section_past_results}
Using max-norm for learning low-rank matrices was pioneered in \cite{srebro2005maximum} where max-norm was used for  collaborative prediction. In this paper we use max-qnorm for tensor completion which is a generalization of a recent result on matrix completion using max-norm constrained optimization \cite{cai2016matrix}. In this section, we review some of the results which is related to M-norm and max-qnorm tensor completion. In particular, we first go over some of the matrix completion results including using nuclear-norm and max-norm and then review some of the results on tensor completion.\newline

Inspired by the result of \cite{fazel2002matrix}, which proved that the nuclear-norm is the convex envelope of rank function, most of the research on matrix completion has focused on using nuclear-norm minimization. Assuming $M$ to be a rank-$r$, $N \times N$ matrix and $M_{\Omega}$ to be the set of $m$ independent samples of this matrix, in \cite{candes2009exact,srebro2004learning}, it was proved that solving
\begin{equation}\label{optimizatin_nuclear_matrix}
\hat{M}:= \text{argmin}\  \|X\|_{\ast} \ \ \ \ \text{subject to} \ \ M_{\Omega}=X_{\Omega},
\end{equation}
recovers the matrix $M$ exactly if $|\Omega| > C N^{1.2} r \log(N)$, provided that the row and column space of the matrix is ``incoherent". This result was later improved in \cite{keshavan2009matrix} to $|\Omega|=O(Nr\log(N))$. There has been significant research in this area since then, either in sharpening the theoretical bound, e.g., \cite{bhojanapalli2014universal,candes2010power,recht2011simpler} or designing efficient algorithms to solve \eqref{optimizatin_nuclear_matrix}, e.g., \cite{jain2013low,cai2010singular}.\newline

More relevant to noisy tensor completion are the results of \cite{candes2010matrix,keshavan2010matrix,cai2016matrix} which consider recovering $M^{\sharp}$ from measurements $Y_{\Omega}$, where $Y=M^{\sharp}+Z$, and $|\Omega|=m$; here $Z$ is a noise matrix. It was proved in \cite{candes2010matrix} that if $\|Z_{\Omega}\|_F \leq \delta$, by solving the nuclear-norm minimization problem
$$\text{argmin} \ \|M\|_{\ast}\ \ \text{subject to}\ \|(X-Y)_{\Omega}\|_F \leq \delta,$$
we can recover $\hat{M}$ where,
$$\frac{1}{N}\|M^{\sharp}-\hat{M}\|_F \leq C\sqrt{\frac{N}{m}} \delta + 2\frac{\delta}{N},$$
provided that there are sufficiently many measurements for perfect recovery in the noiseless case.\newline

Another approach was taken by \cite{keshavan2010matrix} where the authors assume that $\|M^{\sharp}\|_{\infty} \leq \alpha$, and $Z$ is a zero mean random matrix whose entries are i.i.d. with subgaussian-norm $\sigma$. They then suggest initializing the left and right hand singular vectors ($L$ and $R$) from the observations $Y_{\Omega}$ and prove that by solving
$$\underset{L,S,R} {\text{min}}\ \frac{1}{2}\|M^{\sharp}-LSR'\|_F^2\ \ \text{subject to}\ L'L=\mathbb{I}_r,R'R=\mathbb{I}_r,$$
one can recover a rank-$r$ matrix $\hat{M}$ where
$$\frac{1}{N}\|M^{\sharp} - \hat{M}\|_F \leq C \alpha\sqrt{\frac{Nr}{m}} + C'\sigma \sqrt{\frac{Nr\alpha\log(N)}{m}}.$$

Inspired by promising results regarding the use of max-norm for collaborative filtering \cite{srebro2005maximum}, a max-norm constrained optimization was employed in \cite{foygel2011concentration} to solve the noisy matrix completion problem under the uniform sampling assumption. Nuclear norm minimization has been proven to be rate-optimal for matrix completion. However, it is not entirely clear if it is the best approach for non-uniform sampling. In many applications, such as collaborative filtering, the uniform sampling assumption is not a reasonable assumption. For example, in the Netflix problem, some movies get much more attention and therefore have more chance of being rated compared to others. To tackle the issue of non-uniform samples, \cite{negahban2012restricted} suggested using a weighted nuclear norm, imposing probability distributions on samples belonging to each row or column. Due to similar considerations, \cite{cai2016matrix} generalized the max-norm matrix completion to the case of non-uniform sampling and proved that, with high probability, $m=O(\frac{Nr}{\epsilon}\log^3(\frac{1}{\epsilon}))$ samples are sufficient for achieving mean squared recovery error $\epsilon$, where the mean squared error is dependent on the distribution of the observations. To be more precise, in their error bound, indices that have higher probability of being observed are recovered more accurately compared to the entries that have less probability of being observed. In particular, \cite{cai2016matrix} assumed a general sampling distribution as explained in Section \ref{main_result_section} (when $d=2$) that includes both uniform and non-uniform sampling. Assuming that each entry of the noise matrix is a zero mean Gaussian random variable with variance $\sigma$, and $\|M^{\sharp}\|_{\infty} \leq \alpha$, they proved that the solution $\hat{M}_{\m}$ of
$$\underset{\|M\|_{\m}\leq \sqrt{r}\alpha} {\text{min}}\ \|(M^{\sharp}-M)_{\Omega}\|_F^2,$$
assuming $\pi_{\omega} \geq \frac{1}{\mu N^2},\forall \omega \in [N] \times [N]$, satisfies
$$\frac{1}{N^2}\|\hat{M}_{\m}-M^{\sharp}\|_F^2 \leq C \mu (\alpha + \sigma) \alpha \sqrt{\frac{r N}{n}},$$
 with probability greater than $1-2e^{-dN}$. This paper is a generalization of the above result to tensor completion.\newline
 
Finally, we briefly explain some of the past results on tensor completion. To out knowledge, this paper provides the first result that proves linear dependence of the sufficient number of random samples on $N$. It is worth mentioning, though, that \cite{krishnamurthy2013low} proves that $O(N r^{d-0.5} d\log(r))$ adaptively chosen samples is sufficient for exact recovery of tensors. However, the result is heavily dependent on the samples being adaptive.\newline

There is a long list of heuristic algorithms that attempt to solve the tensor completion problem by using different decompositions or matricizations which, in spite of showing good empirical results, are not backed with a theoretical explanation that shows the superiority of using the tensor structure instead of matricization, e.g., see \cite{liu2013tensor,grasedyck2013literature}. The most popular approach is minimizing the sum of nuclear-norm of all the matricizations of the tensor along all modes. To be precise one solves
\begin{equation}\label{SNN}
\underset{X}{\text{min}} \sum_{i=1}^d \beta_i \|X_{(i)}\|_{\ast}\ \text{subject to } X_{\Omega}=T^{\sharp}_{\Omega},
\end{equation}
where $X_{(i)}$ is the mode-$i$ matricization of the tensor (see \cite{liu2013tensor,signoretto2010nuclear,tomioka2010estimation}). The result obtained by solving \eqref{SNN} is highly sensitive on the choice of the weights $\beta_{i}$ and an exact recovery requirement is not available. At least, in the special case of tensor sensing, where the measurements of the tensor are its inner products with random Gaussian tensors, \cite{mu2014square} proves that $m=O(rN^{d-1})$ is necessary for \eqref{SNN}, whereas a more balanced matricization such as $X_{[\floor{\frac{d}{2}}]}$ (as explained in Section \ref{tensor_matricization_section}) can achieve successful recovery with $m=O(r^{\floor{\frac{d}{2}}}N^{\ceil{\frac{d}{2}}})$ Gaussian measurements.\newline

Assuming $T^{\sharp}$ is symmetric and has an orthogonal decomposition, \cite{jain2014provable} proves that when $d=3$, an alternating minimization algorithm can achieve exact recovery from $O(r^5 N^{\frac{3}{2}} \log(N)^4)$ random samples. However, the empirical results of this work show good results for non-symmetric tensors as well if a good initial point can be found.\newline

In \cite{zhang2015exact}, a generalization of the singular value decomposition for tensors, called t-SVD, is used to prove that a third order tensor ($d=3$) can be recovered from $O(r N^2 \log(N)^2)$ measurements, provided that the tensor satisfies some incoherence conditions, called tensor incoherence conditions.\newline

The last related result that we will mention is an interesting theoretical result that generalizes the nuclear-norm to tensors as the dual of the spectral norm and avoids any kind of matricization in the proof \cite{yuan2015tensor}. They show that using the nuclear norm, the sample size requirement for a tensor with low coherence using nuclear-norm is $m=O(\sqrt{rN^d}\log(N))$. Comparing our result with the result of \cite{yuan2015tensor}, an important question that needs to be investigated is whether max-qnorm is a better measure of complexity of low-rank tensors compared to nuclear-norm or whether the difference is just an artifact of the proofs. While we introduce the framework for max-qnorm in this paper, an extensive comparison of these two norms is beyond the scope of this paper. Another difficulty of using tensor nuclear-norm is the lack of sophisticated or even approximate algorithms that can minimize nuclear-norm of a tensor.\newline

We compare our results with some of the above mentioned results in Sections \ref{comparison_past_section} and Section \ref{experiments and algorithms}.
\section{Max-qnorm and atomic M-norm}\label{max_norm_section}
In this section, we introduce the max-qnorm and M-norm of tensors and characterize the unit ball of these norms as tensors that have a specific decomposition with bounded factors. This then helps us to prove a bound on the max-qnorm and M-norm of low-rank tensors that is independent of $N$. The results in this section might be of independent interest and therefore, to give an overview of properties of max-qnorm and M-norm, we state the theorems and some remarks about the theorems and postpone the proofs to Section \ref{section_proofs}.
\subsection{Matrix max-norm}
First, we define the max-norm of matrices which was first defined in \cite{linial2007complexity} as $\gamma_2$ norm. We also mention some of the properties of the matrix max-norm which we generalize later on in this section. Recall that the max-norm of a matrix is defined as
\begin{equation}\label{matrix_maxnorm}
\|M\|_{\text{max}}=\underset{M=U \circ V}{\text{min}}\lbrace \|U\|_{2,\infty}\|V\|_{2,\infty}\rbrace,
\end{equation}
where, $\|U\|_{2,\infty} = \underset{\|x\|_2=1}{\text{sup}} \|Ux\|_{\infty}$ is the maximum $\ell_2$-norm of the rows of $U$ \cite{linial2007complexity,srebro2005rank}.\newline

Considering all the possible factorizations of a matrix $M=U\circ V$, the rank of $M$ is the minimum number of columns in the factors and nuclear-norm of $M$ is the minimum product of the Frobenius norms of the factors. The max-norm, on the other hand, finds the factors with the smallest row-norm as $\|U\|_{2,\infty}$ is the maximum $\ell_2$-norm of the rows of the matrix $U$. Furthermore, it was noticed in \cite{lee2010practical} that max-norm is comparable with nuclear norm in the following sense:
\begin{equation}\label{maxnorm_sum}
\|M\|_{\text{max}} \approx \text{inf}\lbrace\sum_j |\sigma_j|: M=\sum_j \sigma_j u_j v_j^T , \|u_j\|_{\infty}=\|v_j\|_{\infty}=1\rbrace.
\end{equation}
Here, the factor of equivalence is the Grothendieck's constant $K_G \in (1.67,1.79)$. To be precise, $\frac{\text{inf}\sum_j |\sigma_j|}{K_G} \leq \|M\|_{\m} \leq \text{inf}\sum_j |\sigma_j|$, where the infimum is taken over all nuclear decompositions $M=\sum_j \sigma_j u_j v_j^T , \|u_j\|_{\infty}=\|v_j\|_{\infty}=1$.
Moreover, in connection with element-wise $\ell_{\infty}$ norm we have:
\begin{equation}\label{linial}
\|M\|_{\infty} \leq \|M\|_{\text{max}} \leq \sqrt{\text{rank}(M)} \|M\|_{1,\infty} \leq \sqrt{\text{rank}(M)}\|M\|_{\infty}.
\end{equation}
This is an interesting result that shows that we can bound the max-norm of a low-rank matrix by an upper bound that is independent of $N$.
\subsection{Tensor max-qnorm and atomic M-norm}\label{section_tensor_maxnorm_Mnorm}
We generalize the definition of max-norm to tensors as follows. Let $T$ be an order-$d$ tensor. Then
\begin{equation}\label{maxnorm_tensor}
\|T\|_{\text{max}}:= \underset{T=U^{(1)} \circ U^{(2)} \circ \cdots \circ U^{(d)}}{\text{min}}\lbrace \prod_{j=1}^{d} \|U^{(j)}\|_{2,\infty}\rbrace.
\end{equation}
Notice that this definition agrees with the definition of max-norm for matrices when $d=2$. As in the matrix case, the rank of the tensor is the minimum possible number of columns in the low-rank factorization of $T=U^{(1)} \circ U^{(2)} \circ \cdots \circ U^{(d)}$ and the max-qnorm is the minimum row norm of the factors over all such decompositions.\newline

\begin{thm}\label{theorem_max-qnorm_quasi}
For $d \geq 3$, the max-qnorm \eqref{maxnorm_tensor} does not satisfy the triangle inequality. However, it satisfies a quasi-triangle inequality
$$\|X+T\|_{\m}  \leq  2^{\frac{d}{2}-1} (\|X\|_{\m}+\|T\|_{\m}),$$
and, therefore, is a quasi-norm.
\end{thm}

The proof of this theorem is in Section \ref{section_max-qnorm_quasi}. Later on, in Section \ref{section_Max_norm_constrained_LS_estimation}, we prove that a max-qnorm constrained least squares estimation, with the max-qnorm as in \eqref{max_norm_tensor_definition}, breaks the $O(N^{\frac{d}{2}})$ limitation on the number of measurements mainly because of two main properties:
\begin{itemize}
\item Max-qnorm of a bounded low rank tensor does not depend on the size of the tensor.\medskip
\item Defining $T_{\pm}:=\lbrace T \in \{\pm 1\}^{N_1\times N_2 \times \cdots \times N_d}\ |\ \text{rank}(T)=1\rbrace$, the unit ball of the tensor max-qnorm is a subset of $C_d \text{conv}(T_{\pm})$ which is a convex combination of $2^{Nd}$ rank-$1$ sign tensors. Here $C_d$ is a constant that only depends on $d$ and conv($S$) is the convex envelope of the set $S$.
\end{itemize}
However, the max-qnorm in non-convex. To obtain a convex alternative that still satisfies the properties mentioned above, we consider the norm induced by the set $T_{\pm}$ directly; this is an atomic norm as discussed in \cite{chandrasekaran2012convex}. The atomic M-norm of a tensor $T$ is then defined as the gauge of $T_{\pm}$ \cite{rockafellar2015convex} given by
\begin{equation}\label{atomoc_norm_definition}
\|T\|_{M} := \text{inf}\{t>0:T \in t\ \text{conv}(T_{\pm})\}.
\end{equation}
As $T_{\pm}$ is centrally symmetric around the origin and spans $\bigotimes_{j=1}^{d}  \mathbb{R}^{N_j}$, this atomic norm is a convex norm and the gauge function can be rewritten as
\begin{equation}
\|T\|_{M} = \text{inf}\{ \sum_{X \in T_{\pm}} c_X :\ T=\sum_{X \in T_{\pm}} c_X X, c_X \geq 0,X \in T_{\pm}\}.
\end{equation}

\subsection{Unit max-qnorm ball of tensors}
In the next lemma, we prove that, similar to the matrix case, the tensor unit max-qnorm ball is comparable to the set $T_{\pm}$. First define $\mathbb{B}_{\m}^T(1) := \lbrace T \in \mathbb{R}^{N_1 \times \cdots \times N_d}\ |\ \|T\|_{\m} \leq 1\rbrace$ and $\mathbb{B}_{M}(1) := \{ T : \|T\|_{M}\leq 1\}$.
\begin{lem}\label{max_tensor_ball}
The unit ball of the max-qnorm, unit ball of atomic M-norm, and $\text{conv}(T_{\pm})$ satisfy the following:
\begin{enumerate}
\item $\mathbb{B}_{M}(1)=\text{conv}(T_{\pm})$, \medskip
\item $\mathbb{B}_{\m}^T(1)$ $\subset$ $c_1 c_2^d$ conv($T_{\pm}$).
\end{enumerate}
\end{lem}
Here $c_1$ and $c_2$ are derived from the generalized Grothendieck theorem \cite{tonge1978neumann,blei1979multidimensional} which is explained thoroughly in Section \ref{section_max-qnorm_quasi}.\newline

Using Lemma \ref{max_tensor_ball}, it is easy to analyze the Rademacher complexity of the unit ball of these two norms. In fact, noticing that $T_{\pm}$ is a finite class with $|T_{\pm}|<2^{dN}$ and some basic properties of Rademacher complexity we can prove the following lemma. Below, $\hat{R}_S(X)$ denotes the empirical Rademacher complexity of $X$. To keep this section simple, we refer to Section \ref{rademacher_complexity} for the definition of Rademacher complexity and proof of Lemma \ref{lemma_rademacher}.

\begin{lem}\label{lemma_rademacher}
The Rademacher complexity of unit balls of M-norm and max-qnorm is bounded by
 \begin{enumerate}
\item $\underset{S:|S|=m}{sup} \hat{R}_S(\mathbb{B}_{M}(1)) < 6 \sqrt{\frac{dN}{m}}$,\medskip
\item $\underset{S:|S|=m}{sup} \hat{R}_S(\mathbb{B}_{\m}^T(1)) < 6 c_1 c_2^d \sqrt{\frac{dN}{m}}$ .
\end{enumerate}
\end{lem}

\subsection{Max-qnorm and M-norm of bounded low-rank tensors}
Next, we bound the max-qnorm and M-norm of a rank-$r$ tensor whose (entry-wise) infinity norm is less than $\alpha$. First, we bound the max-qnorm and a similar proof can be used to obtain a bound on the M-norm as well which we explain in the Section \ref{section_proof_atomicbound}. As mentioned before, for $d=2$, i.e., the matrix case, an interesting inequality has been proved which does not depend on the size of the matrix, i.e., $\|M\|_{\m} \leq \sqrt{\text{rank}(M)}\ \alpha$. In what follows, we bound the max-qnorm and M-norm of a rank-$r$ tensor $T$ with $\|T\|_{\infty} \leq \alpha$. \newline

\begin{thm}\label{theorem_atomicnorm_bound}
Assume $T \in \mathbb{R}^{N_1 \times \cdots N_d}$ is a rank-$r$ tensor with $\|T\|_{\infty} = \alpha$. Then
\begin{itemize}
\item $\alpha \leq \|T\|_{M} \leq (r\sqrt{r})^{d-1} \alpha.$\medskip
\item $\alpha \leq \|T\|_{\m} \leq \sqrt{r^{d^2-d}} \alpha.$
\end{itemize}
\end{thm}
\noindent
The proofs of these two bounds are similar and both of them can be found in Section \ref{section_proof_atomicbound}. Notice the discrepancy of Theorem \ref{theorem_atomicnorm_bound} when $d=2$. This is an artifact of the proof which hints at the fact that Theorem \ref{theorem_atomicnorm_bound} might be not optimal in $r$ for general $d$ as well.\newline

\section{M-norm constrained tensor completion}\label{TC_maxnorm_section}
In this section, we consider the problem of tensor completion from noisy measurements of a subset of the tensor entries. As explained before, we assume that the indices of the entries that are measured are drawn indepently at random with replacement. Also the tensor of interest is low-rank and has bounded entries. Instead of constraining the problem to the set of low-rank bounded tensors, we consider a more general case and consider the set of bounded tensors with bounded M-norm which includes the set of low-rank bounded tensors. We minimize a constrained least squares (LS) problem given in \eqref{optimization_TC_atomicnorm} below. Similar results can be obtained for a max-qnorm constrained LS. We only provide the final result of the max-qnorm constrained problem in Theorem \ref{theorem_maxqnorm_TC} as the steps are exactly similar to the M-norm constrained one. When $d=2$, i.e., the matrix case, max-norm constrained matrix completion has been thoroughly studied in \cite{cai2016matrix}, so we will not discuss the lemmas and theorems that can be directly used in the tensor case; see \cite{cai2016matrix} for more details.
\subsection{Observation model}\label{section_observation}
Given an order-$d$ tensor $T^{\sharp} \in \mathbb{R}^{N^d}$ and a random subset of indices $S=\{\omega_1,\omega_2,\cdots,\omega_m\},$ $\omega_i \in [N] \times [N] \times \cdots \times [N]$, we observe $m$ noisy entries $\{Y_{\omega_t}\}_{t=1}^{m}$:
\begin{equation}\label{noisy_measurements}
Y_{\omega_t}= T^{\sharp}(\omega_t) + \sigma \xi_t\ , \ \ t=1,\cdots,m,
\end{equation}
for some $\sigma > 0$. The variables $\xi_t$ are zero mean i.i.d. random variables with $\mathbb{E}(\xi_t^2)=1$. The indices in $S$ are drawn randomly with replacement from a predefined probability distribution $\Pi=\{\pi_\omega\}$, for $\omega \in [N]\times[N]\times \cdots\times [N]$, such that $\sum_\omega \pi_\omega=1$. Obviously $\max \pi_\omega\geq \frac{1}{N^d}$. Although it is not a necessary condition for our proof, it is natural to assume that there exist $\mu\geq 1$ such that
\begin{equation*}
 \pi_{\omega} \geq \frac{1}{\mu N^d}\ \ \forall \omega \in [N]^d,
\end{equation*}
which ensures that each entry is observed with some positive probability. This observation model includes both uniform and non-uniform sampling and is a better fit than uniform sampling in many practical applications.
\subsection{M-norm constrained least squares estimation}\label{section_Max_norm_constrained_LS_estimation}
Given a collection of noisy observations $\{Y_{\omega_t}\}_{t=1}^{m}$ of a low-rank tensor $T^{\sharp}$, following the observation model \eqref{noisy_measurements}, we solve a least squares problem to find an estimate of $T^{\sharp}$. Consider the set of bounded M-norm tensors with bounded infinity norm
$$K^T_{M}(\alpha,R):=\lbrace T \in \mathbb{R}^{N_1 \times N_2 \times \cdots \times N_d}: \|T\|_{\infty} \leq \alpha, \|T\|_{M} \leq R \rbrace.$$
Notice that assuming that $T^{\sharp}$ has rank $r$ and $\|T^{\sharp}\|_{\infty} \leq \alpha$, Theorem \ref{theorem_atomicnorm_bound} ensures that a choice of $R = (r\sqrt{r})^{d-1} \alpha$ is sufficient to include $T^{\sharp}$ in $K^T_{M}(\alpha,R)$. Defining 
\begin{equation}\label{L_TC}
\mathcal{L}_{m}(X,Y):=\frac{1}{m}\sum_{t=1}^{m} (X_{\omega_t}-Y_{\omega_t})^2,
\end{equation}
we bound the recovery error for the estimate $\hat{T}_{M}$ obtained by solving the optimization problem
\begin{equation}\label{optimization_TC_atomicnorm}
\hat{T}_{M} = \underset{X}{\text{arg min }} \mathcal{L}_{m}(X,Y) \ \ \ \ \text{subject to}\ \ \ \  X \in K^T_{M}(\alpha,R),\ R\geq \alpha.
\end{equation}
In words, $\hat{T}_{M}$ is a tensor with entries bounded by $\alpha$ and M-norm less than $R$ that is closest to the sampled tensor in Frobenius norm. \newline

We now state the main result on the performance of M-norm constrained tensor completion as in \eqref{optimization_TC_atomicnorm} for recovering a bounded low-rank tensor.
\begin{thm}\label{theorem_atomic_TC}
 Consider an order-$d$ tensor $T^{\sharp} \in \bigotimes_{i=1}^{d} \mathbb{R}^N$ with $\|T^{\sharp}\|_{\infty} \leq \alpha$ and $\|T^{\sharp}\|_{M} \leq R$. Given a collection of noisy observations $\{Y_{\omega_t}\}_{t=1}^{m}$ following the observation model \eqref{noisy_measurements} where the noise sequence $\xi_t$ are i.i.d. standard normal random variables, there exist a constant $C<20$ such that the minimizer $\hat{T}_{M}$ of \eqref{optimization_TC_atomicnorm} satisfies:
 \begin{equation}\label{TC_error}
 \|\hat{T}_{M}-T^{\sharp}\|_{\Pi}^2:=\sum_\omega \pi_\omega (\hat{T}_{M}(\omega)-T^{\sharp}(\omega))^2 \leq C  \left( \sigma(R+\alpha) + R\alpha \right) \sqrt{\frac{dN}{m}},
 \end{equation}
 with probability greater than $1-e^{\frac{-N}{\text{ln}(N)}}-e^{-dN}$. 
\end{thm}
\begin{cor}
If we assume each entry of the tensor is sampled with some positive probability, $\pi_{\omega} \geq \frac{1}{\mu N^d}\ \ \forall \omega \in [N]^d$, then for a sample size $m>dN$, we get
 \begin{equation}\label{TC_frobenious_error}
 \frac{1}{N^d}\|\hat{T}_{M}-T^{\sharp}\|_{F}^2 \leq C \mu  (\alpha + \sigma) R \sqrt{\frac{dN}{m}}.
 \end{equation}
with probability greater than $1- e^{\frac{-N}{\text{ln}(N)}} -e^{-dN}$.
\end{cor}
\begin{rem}
In Section \ref{main_result_section}, we presented a simplified version of the above theorem when $\mu=1$ and $T^{\sharp}$ is a rank-$r$ tensor which uses the bound $\|T^{\sharp}\|_{M} < (r\sqrt{r})^{d-1} \alpha$ proved in Theorem \ref{theorem_atomicnorm_bound}.
\end{rem}
\begin{rem}
The upper bound \eqref{TC_error} is general and does not impose any restrictions on the sampling distribution $\pi$. However, the recovery error depends on the distribution. In particular, the entries that have a bigger probability of being sampled have a better recovery guarantee compared to the ones that are sampled with smaller probability.
\end{rem}
\begin{cor}
Under the same assumptions as in Theorem \ref{theorem_atomic_TC} but assuming instead that $\xi_t$ are independent sub-exponential random variables with sub exponential norm $K$ such that
$$\underset{n=1,\cdots,n}{\max} \mathbb{E} [\exp(\frac{|\xi_t|}{K})] \leq e,$$
for a sample size $m>dN$, we get
 \begin{equation}\label{TC_frobenious_error}
\frac{1}{N^d}\|\hat{T}_{M}-T^{\sharp}\|_{F}^2 \leq C \mu  (\alpha + \sigma K) R \sqrt{\frac{dN}{m}}.
 \end{equation}
with probability greater than $1- 2e^{\frac{-N}{\text{ln}(N)}}$.
\end{cor}

Although equation \eqref{TC_frobenious_error} proves linear dependence of sample complexity with $N$, we are not aware of a polynomial-time method for estimating (or even attempting to estimate) the solution of \eqref{optimization_TC_atomicnorm}. However, we later propose an algorithm that is inspired by max-qnorm constrained tensor completion and illustrate its efficiency numerically. Therefore, now we analyze the error bound of max-qnorm constrained tensor completion which is very similar to the error bound of \eqref{optimization_TC_atomicnorm}. To this end, we define the set of low max-qnorm tensors as
$$K^T_{\m}(\alpha,R):=\lbrace T \in \mathbb{R}^{N_1 \times N_2 \times \cdots \times N_d}: \|T\|_{\infty} \leq \alpha, \|T\|_{\m} \leq R \rbrace.$$
Note that, Theorem \ref{theorem_atomicnorm_bound} ensures that a choice of $R = \sqrt{r^{d^2-d}}\alpha$ is sufficient to include $T^{\sharp}$ in $K^T_{\m}(\alpha,R)$. The following theorem provides the bound on max-qnorm constrained LS estimation.
\begin{thm}\label{theorem_maxqnorm_TC}
 Consider an order-$d$ tensor $T^{\sharp} \in \bigotimes_{i=1}^{d} \mathbb{R}^N$ with $\|T^{\sharp}\|_{\infty} \leq \alpha$ and $\|T^{\sharp}\|_{\m} \leq R$. Given a collection of noisy observations $\{Y_{\omega_t}\}_{t=1}^{m}$ following the observation model \eqref{noisy_measurements} where the noise sequence $\xi_t$ are i.i.d. standard normal random variables, define
\begin{equation}\label{optimization_TC_maxnorm}
\hat{T}_{\m} = \underset{X}{\text{arg min }} \mathcal{L}_{m}(X,Y) \ \ \ \ \text{subject to}\ \ \ \  X \in K^T_{\m}(\alpha,R),\ R\geq \alpha.
\end{equation}
Then there exist a constant $C_d$ such that the minimizer $\hat{T}_{M}$ of \eqref{optimization_TC_atomicnorm} satisfies:
 \begin{equation}\label{TC_qmax_error}
 \|\hat{T}_{\m}-T^{\sharp}\|_{\Pi}^2=\sum_\omega \pi_\omega (\hat{T}_{\m}(\omega)-T^{\sharp}(\omega))^2 \leq C_d  \left( \sigma(R+\alpha) + R\alpha \right) \sqrt{\frac{dN}{m}},
 \end{equation}
 with probability greater than $1-e^{\frac{-N}{\text{ln}(N)}}-e^{-dN}$. 
\end{thm}
\begin{cor}
 Moreover, if we assume each entry of the tensor is sampled with some positive probability, $\pi_{\omega} \geq \frac{1}{\mu N^d}\ \ \forall \omega \in [N]^d$, and for a sample size $m>dN$, we get
 \begin{equation}\label{TC_qfrobenious_error}
 \frac{1}{N^d}\|\hat{T}_{\m}-T^{\sharp}\|_{F}^2 \leq C_d \mu  (\alpha + \sigma) R \sqrt{\frac{dN}{m}}.
 \end{equation}
with probability greater than $1-e^{\frac{-N}{\text{ln}(N)}}-e^{-dN}$.
\end{cor}
\begin{rem}
In Section \ref{main_result_section}, we presented a simplified version of the above theorem when $\mu=1$ and $T^{\sharp}$ is a rank-$r$ tensor which uses the bound $\|T^{\sharp}\|_{\m} < \alpha \sqrt{r^{d^2-d}}$ proved in Theorem \ref{theorem_atomicnorm_bound}.
\end{rem}
The proof of this theorem is very similar to the proof of Theorem \ref{theorem_atomic_TC}. The only differences are: (\RNum{1}) the max-qnorm is a quasi-norm and therefore the max-qnorm of the error tensor ($\hat{T}_{\m} - T^{\sharp}$) is bounded by $2^{d-1} R$; (\RNum{2}) the unit ball of max-qnorm is larger than the unit ball of M-norm. The details of these differences are provided in Remark \ref{remark_proof_maxnorm} in Section \ref{proof_theorem_atomic_TC}.
\section{Comparison to past results}\label{comparison_past_section}
As discussed in Section \ref{section_past_results}, there are several works that have considered max-norm for matrix completion \cite{srebro2005maximum,lee2010practical,foygel2012matrix, shen2014online,fang2015max}. However, the closest work to our result is \cite{cai2016matrix}, where the authors study max-norm constrained matrix completion, which is a special case of max-qnorm constrained tensor completion with $d=2$. Here, we have generalized the framework of \cite{cai2016matrix} to the problem of tensor completion. Although the main ideas of the proof are similar, the new ingredients include building a machinery for analyzing the max-qnorm and M-norm of low rank tensors, as explained in Section \ref{max_norm_section}. As expected, our result reduces to the one in the \cite{cai2016matrix} when $d=2$. More interestingly when $d>2$, compared to the matrix error bound, the only values in upper bound \eqref{TC_error} that change is the upper bound on the max-qnorm of the $d$-th order tensor (which is independent of $N$) and the order $d$, which changes the constants slightly.\newline

As can be seen from Theorem \ref{theorem_atomicnorm_bound}, for a rank-$r$ tensor $T$ with $\|T\|_{\infty} \leq \alpha$, we have $\|T\|_{M} \leq (r\sqrt{r})^{d-1} \alpha$. Therefore, assuming $\alpha=O(1)$, to obtain an error bound of $\frac{1}{N^d}\|\hat{T}_{M}-T^{\sharp}\|_F^2 \leq \epsilon$, it is sufficient to have $m > C \frac{(r\sqrt{r})^{d-1} d N} {\epsilon^2}$ samples. Similarly, using the max-qnorm, for an approximation error bounded by $\epsilon$, it is sufficient to obtain $m > C_d \frac{r^{d^2-d} d N}{\epsilon^2}$ samples. In contrast, the sufficient number of measurements with the best possible matricization is $m> C \frac{r N^{\ceil{\frac{d}{2}}}}{\epsilon^2}$, significantly bigger for higher order tensors.\newline

Tensor completion using nuclear-norm gives significantly inferior bounds as well. In particular, fixing $r$, and $d$, compared to latest results on tensor completion using nuclear-norm \cite{yuan2015tensor}, using M-norm lowers the theoretical sufficient number of measurements from $O(N^{\frac{d}{2}})$ to $O(dN)$.

\section{Information theoretic lower bound}\label{lower_bound_section}
To prove a lower bound on the performance of \eqref{optimization_TC_atomicnorm}, we employ a classical information theoretic technique to establish a minimax lower bound for non-uniform sampling of random tensor completion on the max-qnorm ball. A similar strategy in the matrix case has been used in \cite{davenport20141,cai2016matrix}. In order to derive a lower bound on the performance of \eqref{optimization_TC_atomicnorm}, we find a set of tensors in the set $K^T_{M}$ that are sufficiently far away from each other. Fano's inequality implies that with the finite amount of information that we have, there is no method that can differentiate between all the elements of a set with too many elements and therefore any method will fail to recover at least one of them with a large probability. The main idea and the techniques closely follow \cite[Section 6.2]{cai2016matrix}; therefore we only explain the main steps we take to generalize this approach from matrices to tensors.\newline

For simplicity we assume $N_1=N_2=\cdots=N_d=N$. Similar to the upper bound case, we analyze a general restriction on the max-qnorm of the tensors instead of concentrating on low-rank tensors. Plugging the upper bound of the max-qnorm of low-rank tensors as a special case provides a lower bound for low-rank tensors as well.\newline

Restating the set of bounded low M-norm tensors given by
\begin{equation}\label{K}
K_{M}^T(\alpha,R):=\lbrace T \in \mathbb{R}^{N \times N \times \cdots \times N}: \|T\|_{\infty} \leq \alpha, \|T\|_{M} \leq R \rbrace,
\end{equation}
We will find a lower bound on the recovery error of any method that takes $\{Y_{\omega_t}\}_{t=1}^{m}$ as input and outputs an estimate $\hat{T}$. This includes $\hat{T}_{M}$ that is obtained by
\begin{equation}\label{lower_ptimization_maxnorm}
\hat{T}_{M} = \underset{X}{\text{arg min }} \mathcal{L}_{m}(X,Y) \ \ \ \ \text{subject to}\ \ \ \  X \in K_{M}^T(\alpha,R).
\end{equation}
In particular, we show that when the sampling distribution satisfies
$$\frac{\mu}{N^d} \leq \text{min}_{\omega} \pi_{\omega} \leq \max_{\omega} \pi_{\omega} \leq \frac{L}{N^d},$$
the M-norm constrained least squares estimator is rate optimal on $K_{M}^T(\alpha,R)$.
\begin{thm}\label{theorem_atomic_TC_lowerbound}
Assume that the noise sequence $\xi_t$ are i.i.d. standard normal random variables and the sampling distribution $\Pi$ satisfies $\max_{\omega} \pi_{\omega} \leq \frac{L}{N^d}$. Fix $\alpha$, $R$, and $N$, and $m$ such that
\begin{equation}\label{conditions_lowerbound_theorem}
R^2 \geq \frac{48\alpha^2 K_G^2}{N},
\end{equation}
then the minimax recover error is lower bounded by
\begin{equation}\label{TC_lowerbound}
\underset{\hat{T}_M}{\inf} \underset{T \in K_{M}^T(\alpha,R)}{\sup} \frac{1}{N^d} \mathbb{E}\|\hat{T}_{M}-T\|_F^2 \geq \text{min} \{\frac{\alpha^2}{16},\frac{\sigma R}{128\sqrt{2}K_G}\sqrt{\frac{N}{mL}}\}.
\end{equation}
\end{thm}
\begin{rem}
Comparing the above theorem with \eqref{TC_frobenious_error}, we observe that as long as $\frac{\sigma R}{128\sqrt{2}K_G}\sqrt{\frac{N}{mL}} < \frac{\alpha^2}{16}$, M-norm constrained tensor completion is optimal in both $N$ and $R$.
\end{rem}
\section{Experiments}\label{experiments and algorithms}
In this section, we present algorithms that we use to solve \eqref{optimization_TC_maxnorm} and experiments concerning max-qnorm of specific classes of tensors and max-qnorm constrained tensor completion. As mentioned before most of the typical procedures such as calculating the nuclear norm or even calculating the rank of a tensor are NP-hard. The situation seems even more hopeless if we consider the results of \cite{barak2015noisy} which connects $3$-dimensional tensor completion with refuting $3$-SATs, which has a long line of research behind it. In short, if we assume that either max-qnorm or M-norm is computable in polynomial time, a conjecture of \cite{daniely2013more} for refuting $3$-SATs will be disproved. All these being said, the current paper is the first paper considering max-qnorm for tensor completion and the preliminary results we show in this section are promising, outperforming matricization in every experiment we ran, and even outperforming the TenALS algorithm of \cite{jain2014provable}.\newline

In this section, we concentrate on \eqref{optimization_TC_maxnorm} instead of \eqref{optimization_TC_atomicnorm} as we are not aware of any algorithm that can even attempt to solve \eqref{optimization_TC_atomicnorm} and simple heuristic algorithms we designed for \eqref{optimization_TC_maxnorm} give promising results even though we do not know of any algorithm that is known to converge due to the non-convexity of the optimization problem \eqref{optimization_TC_maxnorm}.\newline

There are two questions that need to be answered while solving \eqref{optimization_TC_maxnorm}. First is how to choose the max-qnorm bound $R$, and second is how to solve the least squares problem once $R$ is fixed. We address both these question in the next sections. We also run some experiments to estimate the tensor max-qnorm of some specific classes of tensors to get an idea of the dependency of the max-qnorm of a tensor on its size and rank. Finally, we compare the results of max-qnorm constrained tensor completion with TenALS and matricizing.


\subsection{Algorithms for max-qnorm constrained least squares estimation}\label{section_algorithms_maxnormTC}
In this section, we introduce a few algorithms that attempt to solve (or approximate the solution of) \eqref{optimization_TC_maxnorm}. Defining $f(V_1, \cdots, V_d,Y):=\mathcal{L}_m((V_1 \circ \cdots \circ V_d),Y)$, we minimize
\begin{equation}\label{optimization_maxnorm_algorithm}
\text{min} f(V_1, \cdots, V_d,Y) \text{ subject to } \max_i (\|V_i\|_{2,\infty}) \leq \sqrt[d]{R},
\end{equation}
where $R$ is the max-qnorm constraint. In the definition of the max-qnorm, there is no limitation on the column size of the factors $V_i$.

In the experiments we run in this section, we limit the factor sizes to $N \times 2N$. Although, this is an arbitrary value and we haven't derived an error bound in the max-qnorm of tensors with this limitation, we believe  (and our experiments also confirm) that this choice is large enough when $r<<N$. We defer the exact details of the effect of this choice on the error bounds to future work.\newline

All the algorithms mentioned in this section are first order methods that are scalable for higher dimensions and just require access to first derivative of the loss function.
\subsubsection{Projected gradient}
The first algorithm is the projected gradient algorithm that for each factor, fixes all the other factors and takes a step according to the gradient of the loss function. Next, we project back all the factors on the set $C:=\{X|\|X\|_{2,\infty} \leq \sqrt[d]{R}\}$. To be precise, for each factor $V_i$, define the matricization of $T=V_1 \circ \cdots \circ V_d$ along the $i$-th dimension, $T_i$, to be $T_i=V_i \circ R_i$ and define $f_i(X):=\mathcal{L}((X \circ R_i),Y_i)$, where $Y_i$ is the matricization of $Y$ along its $i$-th dimension. Fixing a step size $\gamma$, the algorithm updates all the factors in parallel via
\begin{equation}
[V_i] \leftarrow \mathbb{P}_{C}([V_i - \gamma \bigtriangledown(f_i) R_i]).
\end{equation}
where, $\mathbb{P}_C$ simply projects the factor onto the set of matrices with $\ell_2$-infinity norm less than $\sqrt[d]{R}$. This projection looks at each row of the matrix and if the norm of a row is bigger than $\sqrt[d]{R}$, it scales that row back down to $\sqrt[d]{R}$ and leaves other rows unchanged.\newline

This algorithm is a well known algorithm with a lot of efficient implementations and modifications. Furthermore, using armijo line search rule to guarantee sufficient decrease of the loss function, it is guaranteed to find a stationary point of \eqref{optimization_maxnorm_algorithm}.
\subsubsection{Projected quasi-Newton}
Stacking all the factors in a matrix $X$,
$$
X =\left[
\begin{array}{cc}
V_1\\
V_2\\
\vdots\\
V_d
\end{array}\right]
$$
and defining $f(X):=\mathcal{L}_m((V_1 \circ \cdots \circ V_d),Y)$, this algorithm uses BFGS quasi-Newton method to form a quadratic approximation to the function at the current estimate and then uses spectral projected gradient (SPG) method to minimize this quadratic function, constrained to $X \in C$. We use the implementation of \cite{schmidt2009optimizing} which uses limited memory BFGS and uses a Barzilai-Borwein scaling of the gradient, and use a non-monotone Armijo line search along the feasible direction to find the next iterate in the SPG step.
\subsubsection{Stochastic gradient}
The loss function
$$\mathcal{L}_{m}(X,Y)=\frac{1}{m}\sum_{t=1}^{m} (X_{\omega_t}-Y_{\omega_t})^2,$$
is decomposable into the sum of $m$ loss functions, each concerning one observed entry. This makes it very easy to use stochastic gradient methods that at each iteration take one or more of the entries, and find the feasible direction according to this subset of observations. In particular, at each iteration, we take a subset of the $m$ entries, $S \subset \Omega$, and minimize the loss function
$$
\mathcal{L}_{S}(X,Y)=\frac{1}{|S|}\sum_{\omega_t \in S} (X_{\omega_t}-Y_{\omega_t})^2.
$$

This approach is useful when we are dealing with very high dimension sizes and accessing all the measurements at once is not an option or very costly. There has been plenty of research on the efficiency of this method and its recovery guarantees \cite{kushner2012stochastic}. The projection part is done as before with the advantage that we just need to project the rows in the factors that correspond to the subset of entries chosen in this iteration and not necessarily all of them which saves time in large applications.

\subsection{Experiment on max-qnorm of tensors}\label{section_experiments_maxqnorm_tensors}
In this section, we run an experiment to find the dependency of the max-qnorm on its rank and size. To this end, we consider tensors whose low-rank factors come from Gaussian distribution. We also mention the results for tensors coming from random sign factors. Although in comparison with other ways of generating low-rank tensors, these specific classes of tensors do not necessarily represent tensors with highest possibles max-qnorm, they can be helpful in giving us an idea of how does the max-qnorm scale with size and rank.\newline

In order to estimate the max-qnorm of a tensor, we employ a max-qnorm constrained tensor completion while accessing all the entries of the tensor and find the smallest constraint that successfully recovers the tensor. Using the bisection method to estimate the max-qnorm of the tensor, starting from a lower bound and an upper bound for the max-qnorm of the tensor we first check if the tensor can be recovered with max-qnorm bound equal to the average of the upper bound and the lower bound. Next, we increase the lower bound if the max-qnorm constraint is too small for full recovery and reduce the upper bound if the max-qnorm bound is large enough. Algorithm \ref{algorithm_MNfinding} explains this algorithm in more details. For small ranks we get to the approximate max-qnorm very fast, for example, in less than $\log (\text{rank}^{dim-1})+k$ iterations we can estimate the max-qnorm with an error less than $2^{-k}$. Moreover, we assume successful recovery is achieved once the root means squared error (RMSE) is less than a small predefined value. This algorithm becomes faster after the first iteration, as we use the factors found in the previous iteration as a good initial point for the next iteration.\newline

\begin{figure}[h]
\centering
\subfloat[][$3$-dimensional, Gaussian factors]{
\includegraphics[width=0.33\textwidth]{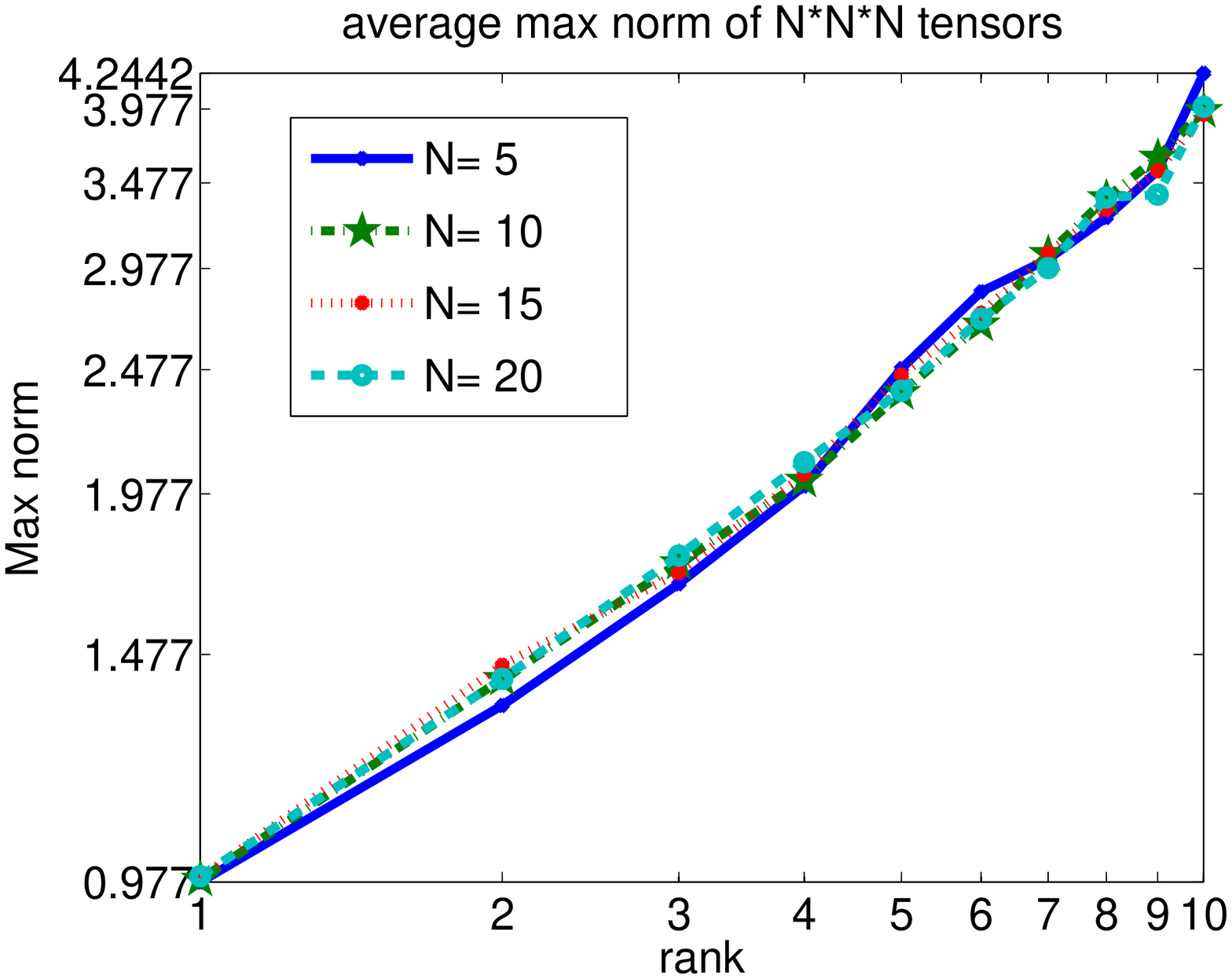}
\label{3dgaussian}}
\subfloat[][$4$-dimensional, Gaussian factors]{
\includegraphics[width=0.33\textwidth]{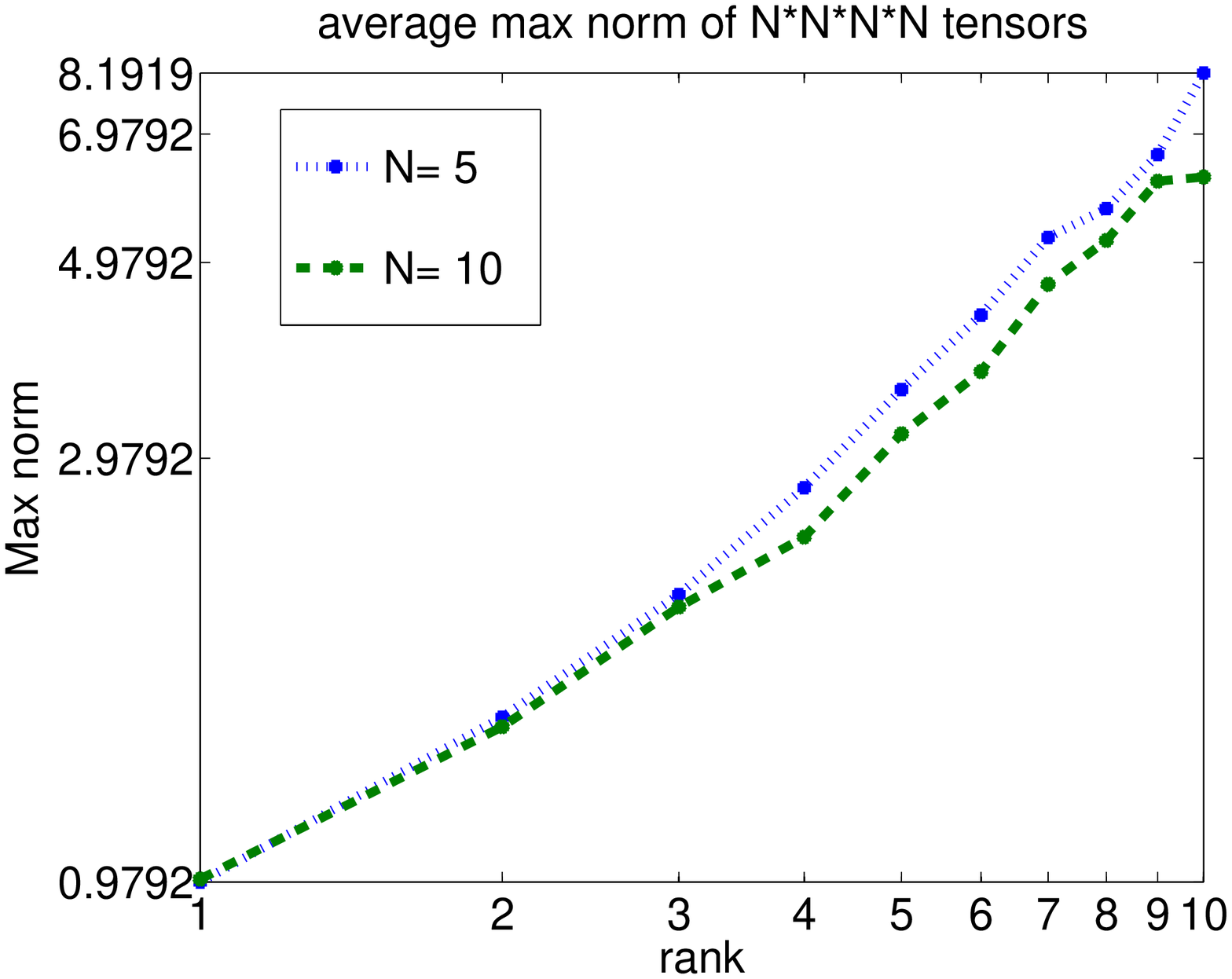}
\label{4dgaussian}}
\subfloat[][$3$ and $4$ dimensional, sign factors]{
\includegraphics[width=0.33\textwidth]{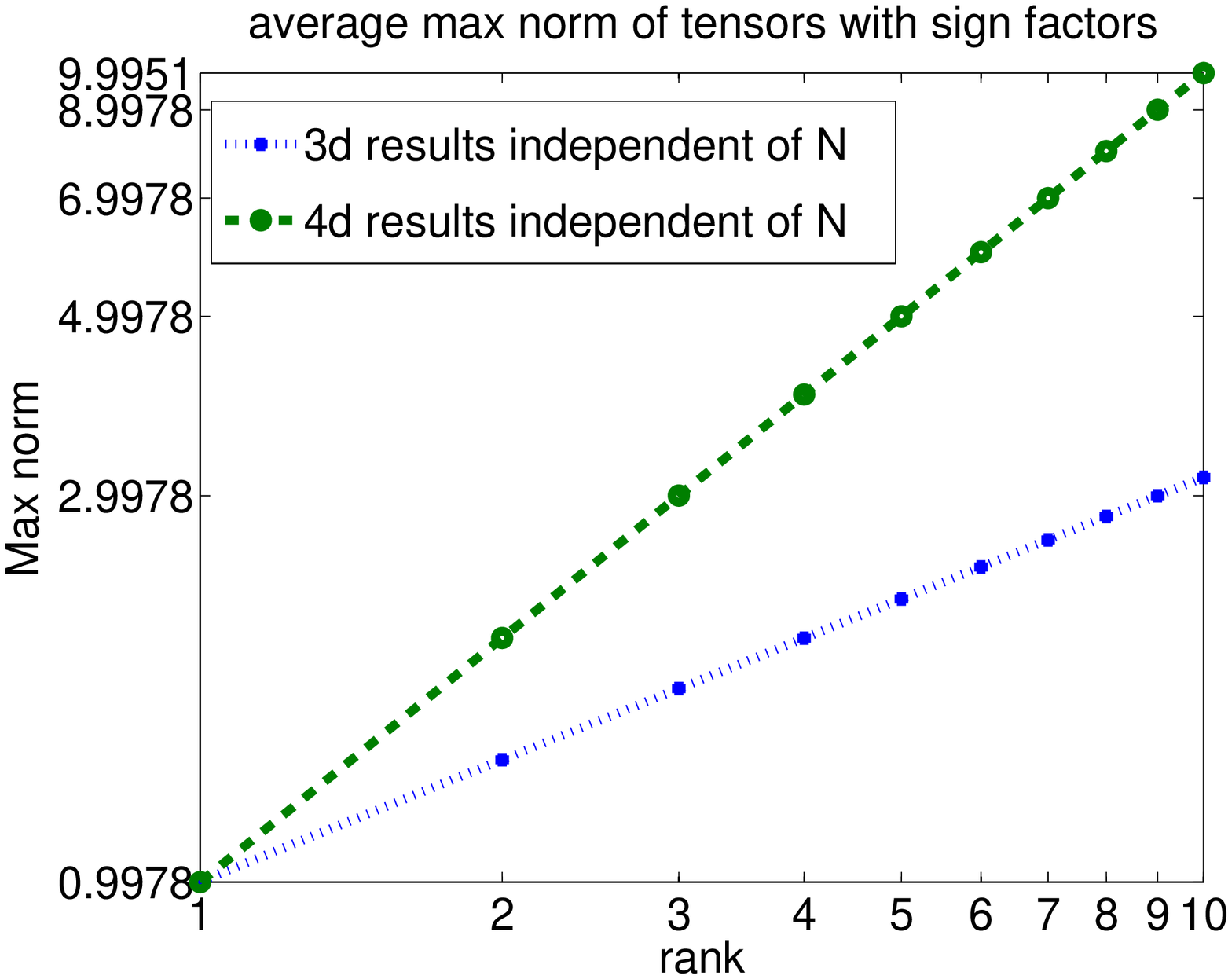}
\label{34dsign}}
\caption{Log-log plot of average max-qnorm of $3$ and $4$ dimensional low-rank tensors obtained by Algorithm \ref{algorithm_MNfinding}, for various rank and sizes, averaged over 15 draws for each rank and size.}
\label{maxnorm_results}
\end{figure}

\begin{algorithm}\caption{Estimating max-qnorm of a tensor $T$}
\begin{algorithmic}[1]\label{algorithm_MNfinding}
\STATE \textbf{Input} $T$, $\Omega=[N_1] \times [N_2] \times \cdots \times [N_d]$, $lowerbound$, $upperbound$
\STATE \textbf{Output} $\|T\|_{\m}$ with an estimation error of at most $0.01$
\FOR{iteration =1 to $\ceil{\log_2 (upperbound-lowerbound)}+6$}
\STATE $\hat{T}=\underset{X \in \mathbb{R}^{N_1} \times \cdots \times \mathbb{R}^{N_d}}{\text{argmin}}$ $\|X_{\Omega}-T_{\Omega}\|_F^2 \text{ subject to } \|X\|_{\m} \leq \frac{lowerbound+upperbound}{2}$
\ENDFOR
\STATE $RMSE=\frac{\|X-T\|_F}{\sqrt{\prod_{i=1}^{i=d}N_i}}$
\IF{$RMSE \leq 1e-3$}
\STATE $upperbound=\frac{lowerbound+upperbound}{2}$
\ELSE 
\STATE $lowerbound=\frac{lowerbound+upperbound}{2}$
\ENDIF
\RETURN $\frac{lowerbound+upperbound}{2}$
\end{algorithmic}
\end{algorithm}\vspace{-0.0in}
Figure \ref{maxnorm_results} shows the results for both $3$ and $4$ dimensional tensors when low-rank factors are drawn either from  Gaussian distribution \ref{maxnorm_results}.a, and \ref{maxnorm_results}.b or from Bernoulli distribution \ref{maxnorm_results}.c. In both cases we have considered $r \in \{1, 2, \cdots, 10\}$, and $N \in \{5,10,15,20\}$ for the $3$-dimensional case and $N \in \{5,10\}$ for the $4$-dimensional case. In all the cases the average max-qnorm is similar for different value of $N$ when rank and order is fixed. These results confirm that max-qnorm of a tensor just depends on its rank and its order and is independent of the size of the tensor.\newline

The results are averaged over 15 experiments. Because of the linear effect of infinity norm of a tensor on its max-qnorm, we rescale all tensors to have$\|T\|_{\infty}=1$ before estimating their max-qnorm. Comparing Figure \ref{maxnorm_results}.a and \ref{maxnorm_results}.b shows that the max-qnorm is around $\sqrt{r}$ for $d=3$, and $r$ for $d=4$. These values are attained exactly when the factors are drawn from Bernoulli random variables. Moreover, the dependence is constant when $d=2$. This suggests a multiplicative increase of $\sqrt{r}$ when the order is increased by one. However, whether or not the actual bound in general case is $O(\sqrt{r^{d^2-d}})$ is an interesting open question.\newline

\subsection{Automatic algorithm for choosing optimal max-qnorm bound $R$}
As explained, before, other than designing algorithms for solving constrained max-qnorm minimization, we need to design a procedure to find good bounds on the max-qnorm. The theoretical bounds found in this paper might not be tight and even if they are proven to be tight, such theoretical upper bounds usually capture the worst-case scenarios which might not be optimal for a general problem. This issue is very important as the result of the tensor completion is very dependent on choosing the right upper bound. Other than this in many practical applications we don't have access to the actual rank of the underlying tensor which shows the importance of finding the upper bound automatically and not as an input to the optimization problem.\newline

\begin{algorithm}\caption{Tensor completion, with cross validation}
\begin{algorithmic}[1]\label{algorithm_TC}
\STATE \textbf{Input} possibly noisy measurements $Y_{\Omega}=T_{\Omega} + \sigma \mathbb{N}(0,1)$, observed entries $\Omega$, $lowerbound$, $upperbound$
\STATE \textbf{Output} $\hat{T}$
\STATE Divide the observations into $\Omega_{\text{train}}$, and $\Omega_{\text{validate}}$
\FOR{iteration =1 to $\ceil{\log_2 (upperbound-lowerbound)}+6$}
\FOR{$iter_{check}$=0 to 4}
\STATE $bound(ietr_{check})= \frac{iter_{check}}{4}*upperbound + \frac{4-iter_{check}}{4}*upperbounds$
\STATE $\hat{T}(ietr_{check})=\underset{X \in \mathbb{R}^{N_1} \times \cdots \times \mathbb{R}^{N_d}}{\text{argmin}}$ $\|X_{\Omega}-Y_{\Omega}\|_F^2 \text{ subject to } \|X\|_{\m} \leq bound_{current} $
\STATE $RMSE(iter_{check})=\frac{\|(\hat{T}(iter_{check})_{\Omega_{\text{validate}}}-(Y)_{\Omega_{\text{validate}}}\|_F}{\sqrt{|\Omega_{\text{validate}|}}}$
\ENDFOR
\STATE $min_{index} = \text{argmin}\ RMSE$
\STATE $lowerbound=bound(min_{index}-1)$
\STATE $upperbound=bound(min_{index}+1)$
\STATE $\hat{T}=\hat{T}(min_{index})$
\ENDFOR
\RETURN $\hat{T}$
\end{algorithmic}
\end{algorithm}\vspace{-0.0in}

The first approach is modifying algorithm \ref{algorithm_MNfinding} to find a good upper bound. There are two complications with generalizing this approach. First, in tensor completion, we don't have access to the full tensor and we have to estimate the recovery error on the indices we have not observed as otherwise choosing a large upper bound can result in over training, i.e, fitting the observations exactly and losing the low-rank (and low max-qnorm) structure of the tensor. The other complication, is that in algorithm \ref{algorithm_MNfinding}, we use $\text{RMSE}<1e-3$ as an approximation for full recovery. Doing such a thing in a noisy problem is not possible and using the RMSE of the noise is an independent problem that depends on the noise-type and is usually not optimal in a general case. Other than this when we have access to the full tensor, using any upper bound bigger than some optimal upper bound results in $\text{RMSE}<1e-3$ which is not the case in tensor completion due to over-fitting. In other words, when we observe the full tensor, over-fitting is meaningless.\newline

To address these two issues, we propose algorithm \ref{algorithm_TC} for tensor completion that uses cross validation for estimating the RMSE and uses a five-point search for the optimal max-qnorm bound.\newline

\begin{figure}[h]
\centering
\includegraphics[width=1\textwidth]{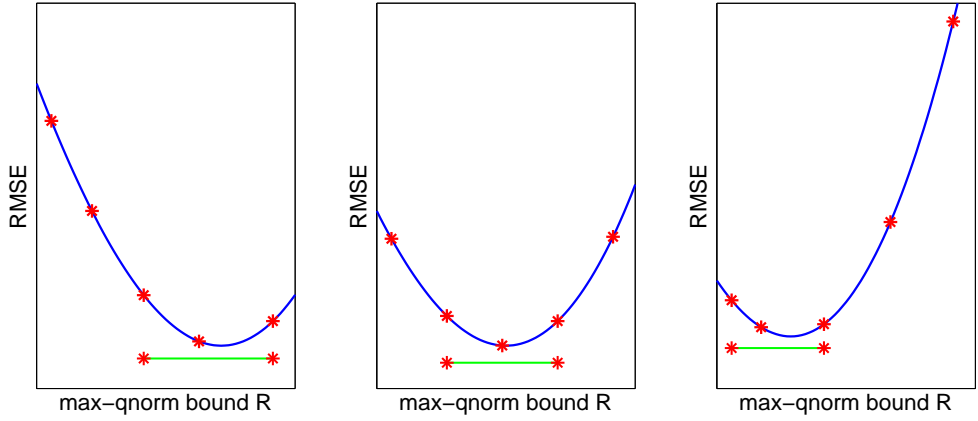}
\caption{All the possible situations in the five-point search algorithm. The leftmost and the rightmost red dots are the previous lower bound and upper bound on the max-qnorm $R$, respectively. The green intervals shows the new lower bound and upper bound based on the RMSE.}
\label{parameter_tuning}
\end{figure}

As explained above we need to find a way to estimate the true RMSE to avoid over-fitting and also as a measure of how good the approximation is. In order to do this, before starting the optimization process, we randomly divide the observed samples, $\Omega$, into two sets: $\Omega_{\text{train}}$ and $\Omega_{\text{validate}}$ and we solve each max-qnorm constrained sub-problem just using the samples in $\Omega_{\text{train}}$ which contains 80\% of the total samples and use the reserved samples $\Omega_{\text{validate}}$ (20\% of total samples) to estimate the RMSE. We are aware of more sophisticated cross validation algorithms that try to remove the bias in the samples as much as possible. However, considering that each max-qnorm constrained tensor completion problem is expensive, our numerical results show that the simple cross validation used in algorithm \ref{algorithm_TC} is good enough for finding an approximately optimal upper bound.\newline

Now we explain algorithm \ref{algorithm_TC} more thoroughly. To find the optimal upper bound we input a large enough upper bound and a small enough lower bound that are bigger and smaller than the optimal max-qnorm bound and iteratively refine these bounds until the two bounds become close to each other. To determine the next upper and lower bounds, checking the middle point is not enough because unlike algorithm \ref{algorithm_MNfinding}, the RMSE is not going to be zero for bounds bigger than the optimal bound. Therefore, other than the lower bound and the upper bound we calculate the RMSE using three points in the interval of the lower bound and the upper bound as well and consider the best bound among those three points to be the center of the new upper bound and the new lower bound. The derivation of this approach is in Algorithm \ref{algorithm_TC}. A hand-wavy reasoning behind this is assume that there is an optimal bound $R^{\sharp}$ that gives the best RMSE, any bound larger than $R^{\sharp}$ results in over training and any bound smaller than $R^{\sharp}$ results in under training. This over-training or under-training becomes more severe when the bound is further from the optimal max-qnorm bound and therefore the problem becomes finding the minimum of a function where we don't know the derivative of the function. Deriving a provably exact algorithm to find the optimal upper bound in such a situation is an interesting question that we postpone to future work. Assuming all the justification above is roughly correct, Figure \ref{parameter_tuning} shows that the five-point method explained in Algorithm \ref{algorithm_TC} finds the optimal upper bound. Considering all the assumptions above to be true, the figure plots the RMSE against the max-qnorm upper bound $R$, and shows all the possible situations that the five points (red stars) can have in such a curve. The green lines shows the new interval bounded by the new lower and upper bound. Notice that the optimal value always stays in the middle of these two bounds. Our numerical experiments in the next section show that although this justification is not mathematically rigorous, the final outcome is very promising. However, providing a better explanation or designing a more rigorous method is an interesting future work.

\subsection{Numerical results of low-rank tensor completion}
In this section we present the results of max-qnorm constrained tensor completion, and compare it with those of matricization and alternating least squares (tenALS) \cite{jain2014provable}. As explained in the beginning of Section \ref{experiments and algorithms}, we pick the low-rank factors to have size $N\times 2N$. Although the choice of $2N$ is arbitrary, we believe it is large enough for small ranks and does not result in large errors. This also has the additional benefit of not requiring the knowledge of the exact rank of the tensors. As explained above, we just assume the knowledge of an upper bound and a lower bound on the max-qnorm of the tensor and use cross validation to the find the optimal max-qnorm bound. Obviously the algorithm becomes faster if these bounds are closer to the actual max-qnorm of the tensor.

%
%

\begin{figure}[h]
\centering
\includegraphics[width=1\textwidth]{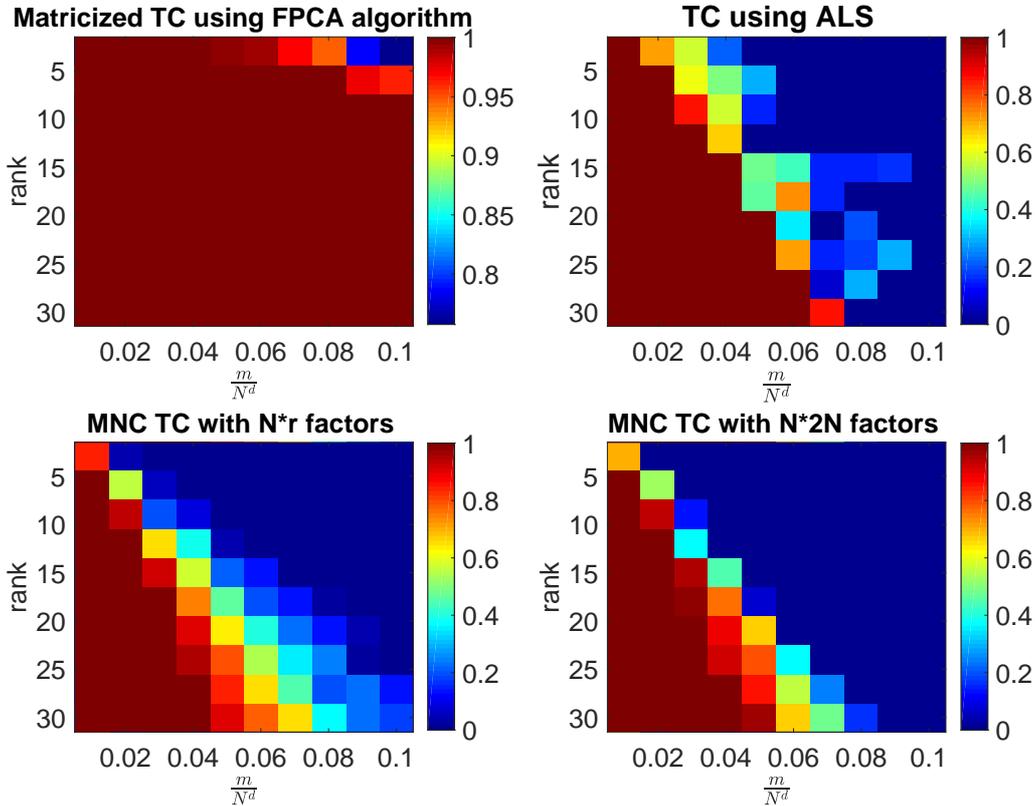}
\caption{{\bf{3-dimensions, no noise }}Average relative recovery error $\frac{\|T_{\text{recovered}} - T^{\sharp}\|_F^2}{\|T^{\sharp}\|_F^2}$ for 3-dimensional tensor $T \in \mathbb{R}^{50\times 50 \times 50}$ and different ranks and samples.}
\label{TC_3d_nonoise}
\end{figure}
\begin{figure}[h]
\centering
\includegraphics[width=1\textwidth]{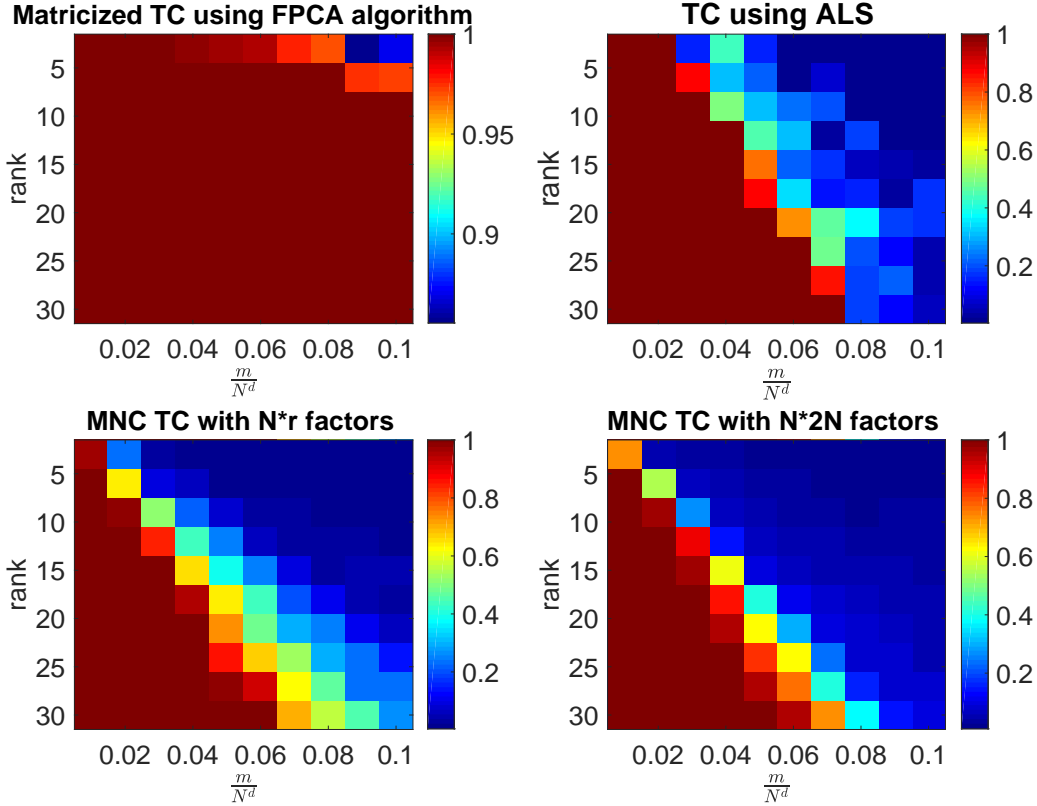}
\caption{{\bf{3-dimensions, 10-db noise }}Average relative recovery error $\frac{\|T_{\text{recovered}} - T^{\sharp}\|_F^2}{\|T^{\sharp}\|_F^2}$ for 3-dimensional tensor $T \in \mathbb{R}^{50\times 50 \times 50}$ and different ranks and samples.}
\label{TC_3d_noisy}
\end{figure}
Figures \ref{TC_3d_nonoise} and \ref{TC_3d_noisy} show the results of completing a $50 \times 50 \times 50$ tensor with $m$ random samples of it taken uniformly at random (without and with 10-db noise respectively). The results are averaged over $15$ experiments, with various ranks ranging from $3$ to $30$ and sampling rate ranging from $0.01$ to $0.1$. The row $i$ and column $j$ of each subplot, shows the average squared relative recovery error $\frac{\|T-\hat{T}\|_F^2}{\|T\|_F^2}$ for a random tensor with rank $i$, where different columns represents different number of samples observed according to \eqref{noisy_measurements}. We rescale the true tensors to have infinity norm equal to $1$ before adding the noise and the RMSE of the noise is around $0.1$. As expected in all experiments we get better results with higher number of measurements $m$, and smaller rank $r$. The matricized results are obtained by applying the Fixed Point Continuation with Approximate SVD algorithm (FPCA), introduced in \cite{ma2011fixed} to flattened $900 \times 30$ matrices (This algorithm results in the best outcomes in our experiments, compared to other noisy matrix completion algorithms). Next, for a more fair comparison we have included the results of tensor completion using alternating least squares (ALS) where the code is provided online \cite{jain2014provable}. The two plots in the second row show the max-qnorm constrained (MNC) results in two scenarios. One with exact low-rank factors and second, as the theory suggests, with factors with larger number of columns. Notice that the exact max-qnorm formulation does not put any limitation on the number of columns of the factors and we chose $2N$ to balance the computational cost and the accuracy of the computed max-qnorms. The results unanimously show the advantage of using $N\times 2N$ factors instead of $N \times r$ factors. Although we are dealing with higher dimensional factors and this makes the algorithm a little slower, using larger factors has the additional benefit of not getting stuck in local minima compared to exact low-rank factors. The ALS algorithm, show some discrepancies in the results, i.e., there are cases that the average error increases with smaller rank or higher sampling rate. We believe this is because of high non convexity in this algorithm. We expected to see similar discrepancies in the max-qnorm results as well but at least for these dimensions and setup, our algorithm seems to be able to run away from local minima which is surprising. We can see the importance of this issue when we compare the results of using $N\times r$ factors and $N \times 2N$ factors. It is worth mentioning here that the ALS algorithm uses a number of initial vectors to use in the initialization step and larger value gives better estimate at the expense of longer processing time. We use 50 vectors so that the time spent by the algorithms is comparable to each other. However, the result can be slightly improved if we use more initial vectors.\newline

The results of matricization is always inferior to those of tensor completion with both ALS and MNC which is expected, especially as here the tensors have an odd order. The difference between matrix completion and MNC with $N \times 2N$ factors is significant. For example, when $\frac{m}{N^d}=0.1$, Max-qnorm constrained TC (MNC) recovers all the tensor with rank less 10, whereas matrix completion starts to fails for ranks bigger than $1$.\newline

\begin{figure}[t]
\centering
\includegraphics[width=1\textwidth]{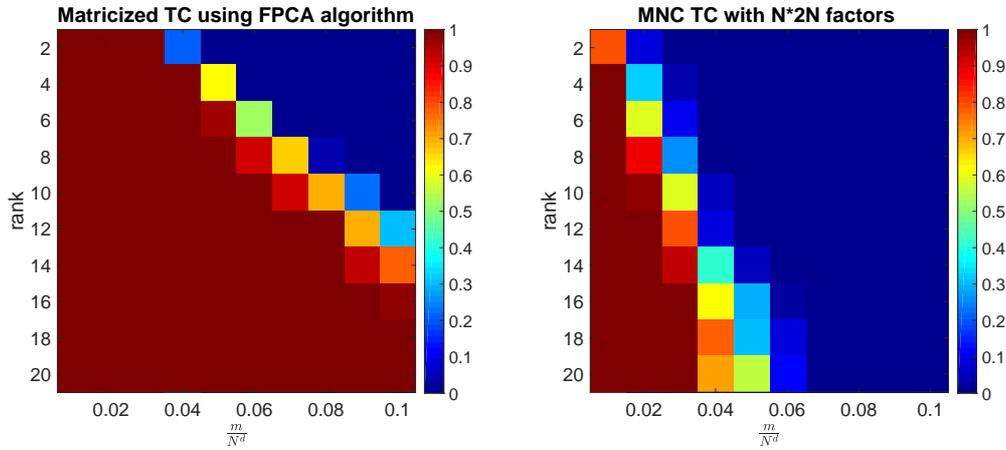}
\caption{{\bf{4-dimensions, no noise }}Average relative recovery error $\frac{\|T_{\text{recovered}} - T^{\sharp}\|_F^2}{\|T^{\sharp}\|_F^2}$ for 4-dimensional tensor $T \in \mathbb{R}^{20\times 20 \times 20 \times 20 \times 20}$ and different ranks and samples. The plot on the left shows the results for the $400 \times 400$ matricized case.}
\label{TC_4d_nonoise}
\end{figure}

\begin{figure}[h]
\centering
\includegraphics[width=1\textwidth]{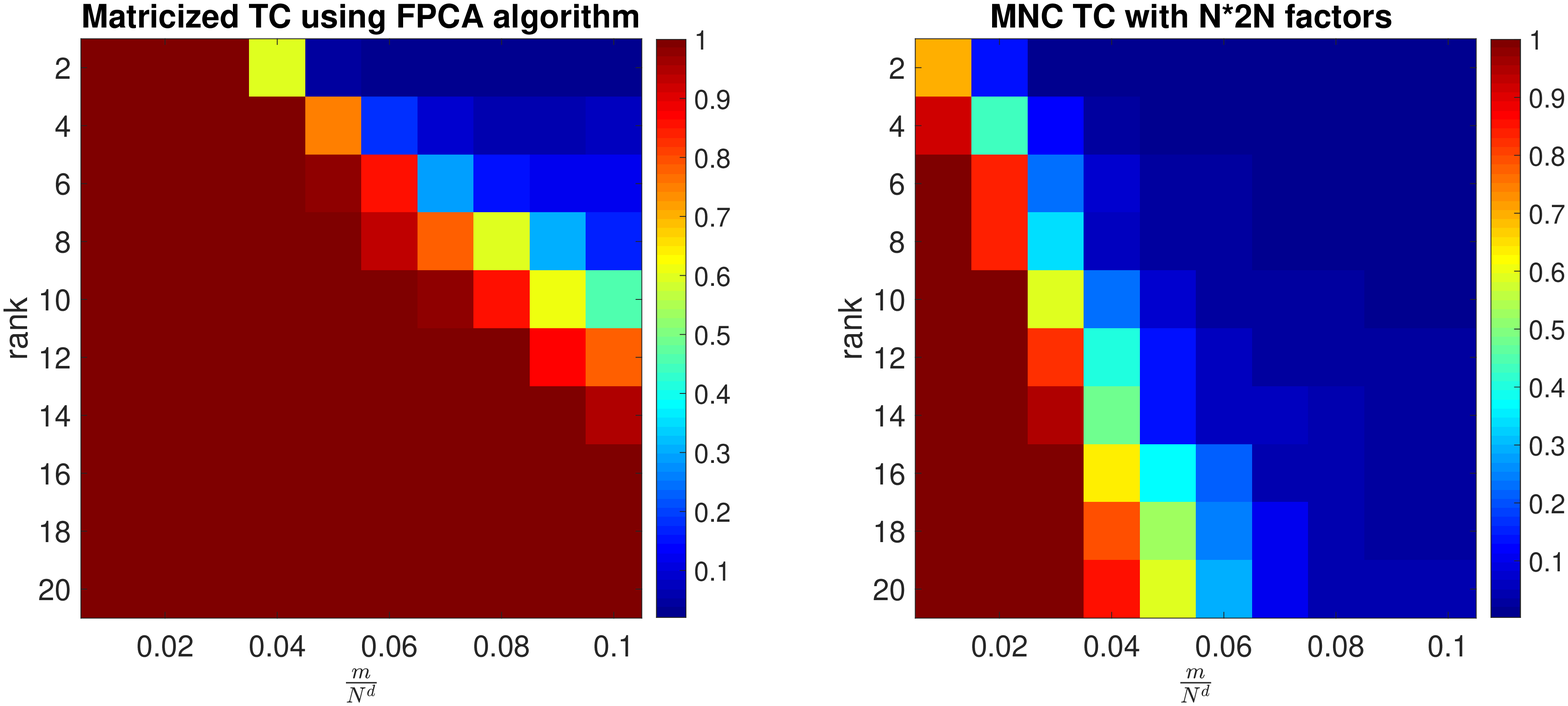}
\caption{{\bf{4-dimensions, 10-db noise }}Average relative recovery error $\frac{\|T_{\text{recovered}} - T^{\sharp}\|_F^2}{\|T^{\sharp}\|_F^2}$ for 4-dimensional tensor $T \in \mathbb{R}^{20\times 20 \times 20 \times 20}$ and different ranks and samples. The plot on the left shows the results for the $400 \times 400$ matricized case.}
\label{TC_4d_noisy}
\end{figure}

In Figures \ref{TC_4d_nonoise} and \ref{TC_4d_noisy} we have included the results for completing a $4$-dimensional $20\times 20 \times 20 \times 20$ tensor. The results are similar to the $3$-dimensional case and therefore we are not including the exact low-rank factor results which are inferior to using $N \times 2N$ factors. The available online code of ALS algorithm is not suitable for $4$-dimensional cases as well. Similar to the $3$-dimensional case MNC TC always beats matrix completion. Notice that because of the even dimension, the matricization can be done in a more balanced way and therefore, results of matrix completion are a little better compared to the $3$-dimensional case but still we can see the huge advantage of using tensor completion instead of matricizing.\newline

It is worth mentioning that although the algorithms take reasonable time for small dimensions used here (less than 15 seconds in each case), it is still slower than matricizing. In other words using tensors enjoys better sample complexity and can be used to save very large data but it comes at the cost of computational complexity. Deriving scalable and efficient algorithms is an interesting and necessary future work for practical applications.
\section{Proofs}\label{section_proofs}
\subsection{Proof of Theorem \ref{theorem_max-qnorm_quasi}}\label{section_max-qnorm_quasi}
Notice that for any tensors $T$ and $X$, there exist decompositions $T=U^{(1)} \circ U^{(2)} \circ \cdots \circ U^{(d)}$ and $X=V^{(1)} \circ V^{(2)} \circ \cdots \circ V^{(d)}$, where $\|U^{(j)}\|_{2,\infty} \leq (\|T\|_{\m})^{\frac{1}{d}}$ and $\|V^{(j)}\|_{2,\infty} \leq (\|X\|_{\m})^{\frac{1}{d}}$. Moreover, one way to factorize the tensor $X+T$ is by concatenating the factors of $X$ and $T$ as $X+T=  [U^{(1)}, V^{(1)}] \circ [U^{(2)}, V^{(2)}] \circ \cdots \circ [U^{(d)}, V^{(d)}]$ and therefore,
\begin{equation*}
\begin{aligned}
\|X+T\|_{\m} &\leq \Pi_{j=1}^d \|[U^{(j)}, V^{(j)}]\|_{2,\infty} \leq \big( \sqrt{\|X\|_{\m}^{\frac{2}{d}}+\|T\|_{\m}^{\frac{2}{d}}} \big)^{d} \leq 2^{\frac{d}{2}-1} (\|X\|_{\m} + \|T\|_{\m}) ,
\end{aligned}
\end{equation*}
which proves that max-qnorm is a quasi-norm. Notice that the last inequality follows from the inequality $|a+b|^p \leq 2^{p-1} (|a|^p + |b|^p)$ for $p>1$. It is easy to check that the max-qnorm satisfies the triangle inequality, when $d=2$. However, this is not true for $d>2$. Next, we prove this for $d=3$ and higher order cases can be proven similarly.\newline
The main challenge in proving that the max-qnorm does not satisfy the triangle-inequality when $d=3$ is that the size of the factors is not fixed. However, it can be observed from the following simple counter-example. Let $T=T_1 + T_2$, where $T_1=\begin{bmatrix} 1\\0 \end{bmatrix} \circ \begin{bmatrix} 1\\0 \end{bmatrix} \circ \begin{bmatrix} 1\\0 \end{bmatrix}$, and $T_2 =\begin{bmatrix} 1\\1 \end{bmatrix} \circ \begin{bmatrix} 1\\1 \end{bmatrix} \circ \begin{bmatrix} 1\\1 \end{bmatrix}$, and note that $T$ is a rank-$2$, $2\times 2 \times 2$ tensor. Here, $T_1$ and $T_2$ are rank-$1$ tensors with $\|T_1\|_{\m}=1$ and $\|T_2\|_{\m}=1$ (notice that for any $T$, $\|T\|_{\m} \geq \|T\|_{\infty}$). Therefore, if max-qnorm satisfies triangle-inequality, then $\|T\|_{\m}$ cannot exceed $2$. In what follows we prove that this is not possible. If $\|T\|_{\m} \leq 2$, then there exists a decomposition $T=U^{(1)} \circ U^{(2)} \circ U^{(3)}$ such that $\|T\|_{\m}=\prod_{j=1}^{3} \|U^{(j)}\|_{2,\infty} \leq 2$, and with a simple rescaling of the factors,
\begin{equation}\label{counterexample_decomposition}
\|U^{(1)}\|_{2,\infty} \leq \sqrt{2},\  \|U^{(2)}\|_{2,\infty} \leq \sqrt{2},\ \|U^{(3)}\|_{2,\infty} \leq 1.
\end{equation}
First, notice that $T$ is an all-ones tensor except for one entry where $T(1,1,1)=2$. Defining the generalized inner product as
\begin{equation}\label{generalized_inner_product}
\langle x_1, \cdots, x_d \rangle := \sum_{i=1}^{k} \prod_{j=1}^{d} x_j(i),
\end{equation}
this means that 
\begin{equation}\label{index_11111}
 \langle U^{(1)}(1,:), U^{(2)}(1,:), U^{(3)}(1,:)\rangle =2.
 \end{equation}
Using Cauchy-Schwarz
\begin{equation}\label{tensor_inner_product_inequality}
\langle U^{(1)}(1,:), U^{(2)}(1,:), U^{(3)}(1,:)\rangle \leq \|U^{(1)}(1,:)\| \ \|U^{(2)}(1,:)\| \ \|U^{(3)}(1,:)\|_{\infty}.
\end{equation}
Combining \eqref{counterexample_decomposition}, \eqref{index_11111}, and \eqref{tensor_inner_product_inequality}, we get
$$2 \leq \|U^{(1)}(1,:)\| \ \|U^{(2)}(1,:)\| \leq \|U^{(1)}\|_{2,\infty} \  \|U^{(2)}\|_{2,\infty} \leq 2,$$
which together with \eqref{counterexample_decomposition} proves that
\begin{equation}\label{U1U2norm}
\|U^{(1)}(1,:)\|=\sqrt{2},\ \text{and}\ \|U^{(2)}(1,:)\|=\sqrt{2}.
\end{equation}
Moreover, similarly
$$2 \leq 2 \|U^{(3)}(1,:)\|_{\infty}\leq 2\ \Rightarrow\ \|U^{(3)}(1,:)\|_{\infty}=1.$$
Notice that $\|U^{(3)}(1,:)\| \leq 1$, and $\|U^{(3)}(1,:)\|_{\infty}=1$ which proves that $U^{(3)}(1,:)$ is an all zeros vector with a single non-zero entry of one. Remember that the number of columns of $U^{(3)}$ is arbitrary. Without loss of generality, we can assume
\begin{equation}
U^{(3)}(1,:)=(1,0,\cdots,0).
\end{equation}
Combining this with \eqref{index_11111}, and \eqref{U1U2norm} we can also prove that
\begin{equation}
U^{(1)}(1,:)=U^{(2)}(1,:)=(\sqrt{2},0,\cdots,0).
\end{equation}
Now from $T(1,1,2)=1$ and the above two equations we have to have $U^{(3)}(2,1)=\frac{1}{2}$ and similarly $U^{(2)}(2,1)=\frac{1}{\sqrt{2}}$. Finally $T(1,2,2)=U^{(1)}(1,1)\ U^{(2)}(2,1)\ U^{(3)}(2,1)=\sqrt{2} \frac{1}{\sqrt{2}} \frac{1}{2} = \frac{1}{2}$ which is a contradiction.\qed

\subsection{Proof of Lemma \ref{max_tensor_ball}}\label{section_max_tensor_ball}
Characterization of the unit ball of the atomic M-norm follows directly from \eqref{atomoc_norm_definition}. By definition, any tensor $T$ with $\|T\|_{M} \leq 1$ is a convex combination of the atoms of $T_{\pm}$, $ T=\sum_{X \in T_{\pm}} c_X X, c_X>0$ with $\sum_{X \in T_{\pm}} c_X=1$. This proves that $\mathbb{B}_{M}(1)=\text{conv}(T_{\pm})$.\newline
 
To characterize the unit ball of max-qnorm, we use a generalization of Grothendieck's theorem to higher order tensors \cite{tonge1978neumann,blei1979multidimensional}. First, we generalize the matrix $\|\cdot\|_{\infty,1}$ norm ($\|M\|_{\infty,1}:=\underset{\|x\|_{\infty}=1}{sup}\|Mx\|_1$) as:
\begin{dfn}\label{definition_inf_one}
$\|T\|_{\infty,1}:=\underset{\|x_1\|_{\infty},\cdots,\|x_d\|_{\infty} \leq 1} {\text{sup}}| \sum_{i_1=1}^{N_1} \cdots \sum_{i_d=1}^{N_d} T(i_1,\cdots,i_d)x_1(i_1)\cdots x_d(i_d)|.$
\end{dfn}
\begin{thm}[{\bf{Generalized Grothendieck theorem}}]\label{bleitong}
Let $T$ be an order-$d$ tensor such that
$$\|T\|_{\infty,1}\ \leftrightarrow \underset{\|x_1\|_{\infty},\cdots,\|x_d\|_{\infty} \leq 1} {\text{sup}}| \sum_{i_1=1}^{N_1} \cdots \sum_{i_d=1}^{N_d} T(i_1,\cdots,i_d)x_1(i_1)\cdots x_d(i_d)| \leq 1,$$
and let $u_{i_j}^j \in \mathbb{R}^k, 1 \leq j \leq d, 1 \leq i_j \leq N_j$ be $\sum N_j$ vectors such that $\|u_{i_j}^j\| \leq 1$. Then
\begin{equation}\label{tensorinf1norm}
| \sum_{i_1=1}^{N_1} \cdots \sum_{i_d=1}^{N_d} T(i_1,\cdots,i_d)\langle u_{i_1}^1, u_{i_2}^2, \cdots, u_{i_d}^d \rangle | \leq c_1 c_2^d,
\end{equation}
where $\langle u_{i_1}^1, u_{i_2}^2, \cdots, u_{i_d}^d \rangle$ is the generalized inner product of $u_{i_1}^1, u_{i_2}^2, \cdots, u_{i_d}^d$ as defined in \eqref{generalized_inner_product}. Here, $c_1 \leq \frac{K_G}{5}$ and $c_2 \leq 2.83$.
\end{thm}
\noindent

Now we use Theorem \ref{bleitong} to prove Lemma \ref{max_tensor_ball}.\newline
\noindent
\textbf{Proof of Lemma \ref{max_tensor_ball}}: The dual norm of the max-qnorm is
\begin{equation}\label{max_dual}
\|T\|_{\m}^{\ast} = \underset {\|U\|_{\m} \leq 1}{\text{max}} \langle T,U \rangle = \underset{\|u_{i_1}^1\|,\cdots,\|u_{i_d}^d\| \leq 1}{\text{max}}\sum_{i_1=1}^{N_1} \cdots \sum_{i_d=1}^{N_d} T(i_1,\cdots,i_d)\langle u_{i_1}^1, u_{i_2}^2, \cdots, u_{i_d}^d \rangle.
\end{equation}
Above, the length of the vectors $u_{i_1}^1,\cdots,u_{i_d}^d$ is not constrained. Using Theorem \ref{bleitong}, $\|T\|_{\m}^{\ast} \leq c_1 c_2^d \|T\|_{\infty,1}$. On the other hand, for $u_{i_1}^1,\cdots,u_{i_d}^d \in \mathbb{R}$ the right hand side of \eqref{max_dual} is equal to $\|T\|_{\infty,1}$. Therefore, $\|T\|_{\infty,1} \leq \|T\|_{\m}^{\ast}$. Taking the dual:
\begin{equation}
\frac{\|T\|_{\infty,1}^{\ast}}{c_1 c_2^d} \leq (\|T\|_{\m}^{\ast})^{\ast} \leq \|T\|_{\infty,1}^{\ast}
\end{equation}
Notice that the max-qnorm, defined in \eqref{max_norm_tensor_definition} is a quasi-norm and therefore, $(\|T\|_{\m}^{\ast})^{\ast}$ is not equal to $\|T\|_{\m}$. However, notice that the max-qnorm is absolutely homogeneous and therefore,
$$(\|T\|_{\m}^{\ast})^{\ast} = \underset{\|Z\|_{\m}^{\ast}\leq 1}{\m} \langle T,Z \rangle \leq \|T\|_{\m}.$$
which implies that
\begin{equation}\label{max_inequality}
\frac{\|T\|_{\infty,1}^{\ast}}{c_1 c_2^d} \leq \|T\|_{\m}.
\end{equation}
To calculate the unit ball of $\|.\|_{\infty,1}^{\ast}$, notice that the argument of the supremum in Definition \ref{definition_inf_one} is linear in each variable $x_j(i_j)$ and as $-1 \leq x_j(i_j) \leq 1$, the supremum is achieved when $x_j(i_j)=\pm 1$ which means that $\|T\|_{\infty,1}=\underset{U \in T_{\pm}}{sup} |\langle T,U \rangle|$. Therefore, $\text{conv}(T_{\pm})$ is the unit ball of $\|.\|_{\infty,1}^{\ast}$ and Lemma \ref{max_tensor_ball} (\RNum{2}) follows from \eqref{max_inequality}.

\subsection{Proof of Theorem \ref{theorem_atomicnorm_bound}}\label{section_proof_atomicbound}
In order to prove the tensor max-qnorm bound, we first sketch the proof of \cite{rashtchian2016bounded} for the matrix case. That is, assuming that $M$ ia s matrix with $\text{rank}(M)=r$ and $\|M\|_{\infty}\leq \alpha$, we show that there exist a decomposition $M=U \circ V$ where $U \in \mathbb{R}^{N_1 \times r}, V \in \mathbb{R}^{N_2 \times r}$ and $\|U\|_{2,\infty} \leq \sqrt{r}, \|V\|_{2,\infty} \leq \alpha$. To prove this, we first state a version of the John's theorem \cite{rashtchian2016bounded}.
\begin{thm}[John's theorem \cite{john2014extremum}]
For any full-dimensional symmetric convex set $K \subseteq \mathbb{R}^r$ and any ellipsoid $E \subseteq \mathbb{R}^r$
that is centered at the origin, there exists an invertible linear map $S$ so that $E \subseteq S(K) \subseteq \sqrt{r} E$.
\end{thm}
\begin{thm}\label{john_matrix}\cite[Corollary 2.2]{rashtchian2016bounded}
For any rank-$r$ matrix $M \in \mathbb{R}^{N_1 \times N_2}$ with $\|M\|_{\infty}\leq \alpha$ there exist vectors $u_1,\cdots,u_{N_1},v_1,\cdots,v_{N_2} \in \mathbb{R}^r$ such that $\langle u_i,v_j\rangle =M_{i,j}$ and $\|u_i\|\leq \sqrt{r}$ and $\|v_j\|\leq \alpha$.
\end{thm}

The proof is based on considering any rank-$r$ decomposition of $M=X\circ Y$ where, $X \in \mathbb{R}^{N_1 \times r}$ and $Y \in \mathbb{R}^{N_2 \times r}$ and $M_{i,j}=\langle x_i,y_j\rangle$. Defining $K$ to be the convex hull of the set $\{ \pm x_i : i \in [N_1]\}$. Then using the linear map $S$ in John's Theorem for the set $K$ with the ellipsoid $E=\mathbb{B}_r:=\{x \in \mathbb{R}^r : \|x\|_2 \leq 1\}$, the decomposition $M=(XS)\circ (YS^{-1})$ satisfies the conditions of Theorem \ref{john_matrix} \cite{rashtchian2016bounded}.\newline

The following lemma proves the existence of a nuclear decomposition for bounded rank-$r$ tensors, which can be used directly to bound the M-norm of a bounded rank-$r$ tensor.
\begin{lem}\label{inductive_atomicnorm_decom}
Any order-$d$, rank-$r$ tensor $T$, with $\|T\|_{\infty} \leq \alpha$ can be decomposed into $r^{d-1}$ rank-one tensors whose components have unit infinity norm such that
\begin{equation}
T=\sum_{j=1}^{r^{d-1}} \sigma_j u_j^1 \circ u_j^2 \circ \cdots \circ u_j^d ,\  \|u_j^1\|_{\infty},\cdots,\|u_j^d\|_{\infty} \leq 1, \ \text{with } \sum |\sigma_j| \leq (r\sqrt{r})^{d-1} \alpha.
\end{equation}
\end{lem}
{\bf{Proof}}: We prove this lemma by induction. The proof for $d=2$ follows directly from applying John's theorem to a rank-$r$ decomposition of $T$, i.e., $T=XS \circ YS^{-1}$ where $T=X \circ Y$. Now assume an order-$d$ tensor which can be written as $T=\sum_{j=1}^{r} \lambda_j v_j^1 \circ v_j^2 \circ \cdots \circ v_j^d$ and $\|T\|_{\infty} \leq \alpha$. Matricizing along the first dimension results in $T_{[1]} = \sum_{i=1}^r (\lambda_i u_i^{(1)})  \circ (u_i^{(2)} \otimes \cdots \otimes u_i^{(d)})$. Using MATLAB notation we can write $T_{[1]}=U \circ V$ where $U(:,i)=\lambda_i u_i^{(1)} \in \mathbb{R}^{N_1}$, and $V(:,i)=u_i^{(2)} \otimes \cdots \otimes u_i^{(d)} \in \mathbb{R}^{\Pi_{k=2}^{k=d}N_k}$.\newline
Using John's theorem, there exist $S \in \mathbb{R}^{r \times r}$ where $T_{[1]}=X \circ Y$ where $X=US$, $Y=VS^{-1}$, $\|X\|_{\infty} \leq \|X\|_{2,\infty} \leq \sqrt{r}$, and  $\|Y\|_{\infty} \leq \|Y\|_{2,\infty} \leq \alpha$. Furthermore, each column of $Y$ is a linear combination of the columns of $V$, i.e., there exist $\zeta_{1}, \cdots \zeta_{r}$ such that $Y(:,i)=\sum_{j=1}^r \zeta_{j} (u_j^{(2)} \otimes \cdots \otimes u_j^{(d)})$. Therefore, unfolding $i$-th column of $Y$ into a $(d-1)$-dimensional tensor $E_i \in \mathbb{R}^{N_2 \times \cdots \times N_d}$ would result in a rank-$r$, $(d-1)$-dimensional tensor with $\|E_i\|_{\infty} \leq \|Y\|_{\infty} \leq \alpha$. By induction, $E_i$ can be decomposed into $r^{d-2}$ rank-one tensors with bounded factors, i.e., $E_i =\sum_{j=1}^{r^{d-2}} \sigma_{i,j} v_{i,j}^1 \circ v_{i,j}^2 \circ \cdots \circ v_{i,j}^d$, where $\|v_{i,j}\|_{\infty} \leq 1$ and $\sum |\sigma_{i,j}| \leq (r\sqrt{r})^{d-2} \alpha $.\newline

Going back to the original tensor, as $T_{[1]}= X \circ Y$, we also have $T= \sum_{i=1}^r X(:,i) \circ (\sum_{j=1}^{r^{d-2}} \sigma_{i,j} v_{i,j}^1 \circ v_{i,j}^2 \circ \cdots \circ v_{i,j}^d)$. Notice that $\|X(:,i)\|_{\infty} \leq \sqrt{r}$. Therefore, by distributing the outer product and rearranging, we get $T=\sum_{j=1}^{r^{d-1}} \sigma_j u_j^1 \circ u_j^2 \circ \cdots \circ u_j^d ,\ \|u_j^1\|_{\infty},\cdots,\|u_j^d\|_{\infty} \leq 1$ and $\sum |\sigma_j| \leq \sum_{i=1}^r \sqrt{r} \left( (r\sqrt{r})^{d-2} \alpha \right) = (r\sqrt{r})^{d-1} \alpha$, which concludes the proof or Lemma \ref{inductive_atomicnorm_decom}. This lemma can be used directly to bound the M-norm of a bounded rank-$r$ tensor.\newline

Next, we bound the max-qnorm of a bounded rank-$r$ tensor. The following lemma proves the existence of a nuclear decomposition for bounded rank-$r$ tensors, which can be used directly to bound their max-qnorm. As the max norm is linear, without loss of generality we assume $\|T\|_{\infty} \leq 1$.
\begin{lem}\label{inductive_maxnorm_decom}
Any order-$d$, rank-$r$ tensor $T \in \bigotimes_{i=1}^{d}  \mathbb{R}^{N_i}$, with $\|T\|_{\infty} \leq 1$ can be decomposed into $r^{d-1}$ rank-one tensors, $T=\sum_{j=1}^{r^{d-1}} u_j^1 \circ u_j^2 \circ \cdots \circ u_j^d$, where:
\begin{equation}
\sum_{j=1}^{r^{d-1}} (u_j^k(t))^2 \leq r^{d-1}\ \text{for any } 1\leq k \leq d,\ 1\leq t\leq N_k.
\end{equation}

Notice that $\sqrt{\sum_{j=1}^{r^{d-1}} (u_j^k(t))^2}$ is the spectral norm of $j$-th row of $k$-th factor of $T$, i.e., $\sum_{j=1}^{r^{d-1}} (u_j^k(t))^2 \leq r^{d-1}$ means that the two-infinity norm of the factors is bounded by $\sqrt{r^{d-1}}$.
\end{lem}
\begin{rem}{[Proof by Lemma \ref{inductive_atomicnorm_decom}]}
At the end of this subsection, we provide a proof for the lemma as stated above. However, using the decomposition obtained in Lemma \ref{inductive_atomicnorm_decom}, we can find a decomposition with $\sum_{j=1}^{r^{d-1}} (u_j^k(t))^2 \leq r^{d}$. To do this notice that by Lemma \ref{inductive_atomicnorm_decom}, and defining $\vec{\sigma}:=\{\sigma_1, \cdots,\sigma_{r^{d-1}}\}$ we can write
$$T=\sum_{j=1}^{r^{d-1}} \sigma_j v_j^1 \circ v_j^2 \circ \cdots \circ v_j^d ,\  \|v_j^1\|_{\infty},\cdots,\|v_j^d\|_{\infty} \leq 1, \ \text{with } \|\vec{\sigma}\|_1\leq (r\sqrt{r})^{d-1}.$$
Now define
$$u_j^k:=(\sigma_j)^{\frac{1}{d}} v_j^k\ \text{for any } 1\leq k \leq d,\ 1\leq t\leq N_k.$$
It is easy to check that $T=\sum_{j=1}^{r^{d-1}} u_j^1 \circ u_j^2 \circ \cdots \circ u_j^d$ and
$$\sum_{j=1}^{r^{d-1}} (u_j^k(t))^2 = \sum_{j=1}^{r^{d-1}} \sigma_j^{\frac{2}{d}} (v_j^k(t))^2 \leq \sum_{j=1}^{r^{d-1}} \sigma_j^{\frac{2}{d}} = \|\vec{\sigma}\|_{\frac{2}{d}}^{\frac{2}{d}}.$$
Using Holder's inequality, when $d \geq 2$ we have
$$\sum_{j=1}^{r^{d-1}} (u_j^k(t))^2 = \|\vec{\sigma}\|_{\frac{2}{d}}^{\frac{2}{d}} \leq \| \vec{\sigma} \|_1^{\frac{2}{d}} (r^{d-1})^{1-\frac{2}{d}} \leq r^{\frac{3d-3}{d}} r^{\frac{(d-1)(d-2)}{d}}=r^{\frac{(d-1)(d+1)}{d}} \leq r^d$$
This proves an upper bound which is close to the one in the lemma. To get a more optimal upper bound (the one stated in the Lemma \ref{inductive_maxnorm_decom}) we need to go over the induction steps as explained below.
\end{rem}
\noindent
{\bf{Proof of Lemma \ref{inductive_maxnorm_decom}}}: We prove this lemma by induction. The proof for $d=2$ follows directly from applying John's theorem to a rank-$r$ decomposition of $T$, i.e., $T=XS \circ YS^{-1}$ where $T=X \circ Y$. Now assume an order-$d$ tensor which can be written as $T=\sum_{j=1}^{r} u_j^1 \circ u_j^2 \circ \cdots \circ u_j^d$ and $\|T\|_{\infty} \leq 1$. Matricizing along the first dimension results in $T_{[1]} = \sum_{i=1}^r (u_i^{(1)})  \circ (u_i^{(2)} \otimes \cdots \otimes u_i^{(d)})$. Using matrix notation we can write $T_{[1]}=U \circ V$ where $U(:,i)= u_i^{(1)} \in \mathbb{R}^{N_1}$, and $V(:,i)=u_i^{(2)} \otimes \cdots \otimes u_i^{(d)} \in \mathbb{R}^{\Pi_{k=2}^{k=d}N_k}$.\newline

Using John's theorem, there exist $S \in \mathbb{R}^{r \times r}$ where $T_{[1]}=X \circ Y$ where $X=US$, $Y=VS^{-1}$, $\|X\|_{2,\infty} \leq \sqrt{r}$, and  $\|Y\|_{\infty} \leq \|Y\|_{2,\infty} \leq 1$. More importantly, each column of $Y$ is a linear combination of the columns of  $V$. More precisely, there exist $\zeta_{1}, \cdots \zeta_{r}$ such that $Y(:,i)=\sum_{j=1}^r \zeta_{j} (u_j^{(2)} \otimes \cdots \otimes u_j^{(d)})$. Therefore, unfolding $i$-th column of $Y$ into a $(d-1)$-dimensional tensor $E_i \in \mathbb{R}^{N_2 \times \cdots \times N_d}$ would result in a rank-$r$, $(d-1)$-dimensional tensor with $\|E_i\|_{\infty} \leq \|Y\|_{\infty} \leq 1$. By induction, $E_i$ can be decomposed into $r^{d-2}$ rank-one tensors where $E_i =\sum_{j=1}^{r^{d-2}} v_{i,j}^2 \circ v_{i,j}^3 \circ \cdots \circ v_{i,j}^d$, where $\sum_{j=1}^{r^{d-2}} (v_{i,j}^k(t))^2 \leq r^{d-2}$ for any $2\leq k\leq d$ and any $1\leq t \leq N_k$. Notice that the factors start from $v_{i,j}^2$ to emphasize that $E$ is generated from the indices in the dimensions $2$ to $d$.\newline

Going back to the original tensor, as $T_{[1]}= X \circ Y$, we can write
$$T= \sum_{i=1}^r X(:,i) \circ (\sum_{j=1}^{r^{d-2}} v_{i,j}^2 \circ v_{i,j}^3 \circ \cdots \circ v_{i,j}^d).$$

By distributing the outer product we get $T= \sum_{i=1}^r \sum_{j=1}^{r^{d-2}} X(:,i) \circ v_{i,j}^2 \circ v_{i,j}^3 \circ \cdots \circ v_{i,j}^d$. Renaming the vectors in the factors we get
$$T = \sum_{k=1}^{r^{d-1}} u_k^1 \circ u_k^2 \circ \cdots \circ u_k^d.$$

Now we bound the max norm of $T$ using this decomposition by considering each factor separately using the information we have about $X$ and $E_i$s.\newline

Starting from the first factor, notice that $\|X\|_{2,\infty} \leq \sqrt{r}$ or more precisely $\sum_{i=1}^r X(t,i)^2 \leq r$ for any $1 \leq t \leq N_1$. By Careful examining of the two decompositions of $T$ stated above, we get
$$u_k^1=X(:,\text{mod}(k-1,r)+1)$$
and therefore
\begin{equation}\label{square_sum_first_factor}
\sum_{k=1}^{r^{d-1}} (u_k^1(t))^2= r^{d-2} \sum_{i=1}^{r} X(t,i)^2 \leq r^{d-2} r=r^{d-1},\ \text{for any } 1 \leq t \leq N_1,
\end{equation}
which proves the lemma for the vectors in the first dimension of the decomposition.\newline

For the second dimension, define $j:=\text{mod}(k-1,r^{d-2})+1$, and $j:=\frac{k-j}{r^{d-2}}+1$. Then
$$u_k^2=v_{i,j}^k,$$
and therefore,
\begin{equation}\label{square_sum_second_factor}
\sum_{k=1}^{r^{d-1}} (u_k^2(t))^2= \sum_{i=1}^r \sum_{j=1}^{r^{d-2}} (v_{i,j}^2(t))^2 \leq \sum_{i=1}^ r r^{d-2} \leq r^{d-1},\ \text{for any } 1 \leq t \leq N_2,
\end{equation}
which finishes the proof the lemma for the vectors in the second dimension. All the other dimensions can be bounded in an exact similar way to the second dimension.\newline

The bound on the max-qnorm of a bounded rank-$r$ tensor, follows directly from Lemma \ref{inductive_maxnorm_decom} and definition of tensor max-qnorm.
\begin{rem}
In both lemmas \ref{inductive_atomicnorm_decom} and \ref{inductive_maxnorm_decom}, we start by decomposing a tensor $T=U_1 \circ U_2 \circ \cdots \circ U_d$ into $T_{[1]}=U_1 \circ V$ and generating $K$ (in the John's theorem) by the rows of the factor $U_1$. Notice that John's theorem requires the set $K$ to be full-dimensional. This condition is satisfied in the matrix case as the low rank decomposition of a matrix (with the smallest rank) is full-dimensional. However, this is not necessarily the case for tensors. In other words, the matricization along a dimension might have smaller rank than the original tensor. To take care of this issue, consider a factor $U_{add}$ with the same size of $U_1$ such that $U_1+U_{add}$ is full-dimensional. Now the tensor $T_{\epsilon} = T +  \epsilon U_{add} \circ U_2 \circ \cdots \circ U_d\epsilon$ satisfies the conditions of the John's theorem and by taking $\epsilon$ to zero we can prove that $\|T\|_M = \|T_{\epsilon}\|_M$ and $\|T\|_{\m} = \|T_{\epsilon}\|_{\m}$. Notice that M-norm is convex and max-qnorm satisfies $\|X+T\|_{\m} \leq \big( \sqrt{\|X\|_{\m}^{\frac{2}{d}} + \|T\|_{\m}^{\frac{2}{d}} } \big)^d$.
\end{rem}

\subsection{Proof of Theorem \ref{theorem_atomic_TC}}\label{proof_theorem_atomic_TC}
In this section, we prove Theorem \ref{theorem_atomic_TC}. We make use of Lemma \ref{lemma_rademacher} and Theorem \ref{theorem_atomicnorm_bound} repeatedly. However, some parts of the other calculations are simple manipulations of the proof in \cite[Section 6]{cai2016matrix}.\newline

For ease of notation define $\hat{T}:=\hat{T}_{M}$. Notice that $T^{\sharp}$ is feasible for \eqref{optimization_TC_atomicnorm} and therefore,
\begin{equation*}
\frac{1}{m}\sum_{t=1}^{m} (\hat{T}_{\omega_t}-Y_{\omega_t})^2 \leq \frac{1}{m}\sum_{t=1}^{m} (T^{\sharp}_{\omega_t}-Y_{\omega_t})^2.
\end{equation*}

Plugging in $Y_{\omega_t}= T(\omega_t) + \sigma \xi_t$ and defining $\Delta=\hat{T}-T^{\sharp} \in K_{M}^T(2\alpha,2R)$ we get
\begin{equation}\label{delta_TC_equation}
\frac{1}{m}\sum_{t=1}^{m} \Delta(\omega_t)^2 \leq \frac{2\sigma}{m}\sum_{t=1}^{m} \xi_t \Delta(\omega_t).
\end{equation}

The proof is based on a lower bound on the left hand side of \eqref{delta_TC_equation} and an upper bound on its right hand side.
\subsubsection{Upper bound on right hand side of \eqref{delta_TC_equation}}
First, we bound $\hat{R}_m(\alpha,R) := \underset{T \in K_M^T(\alpha,R)} {\sup} |\frac{1}{m} \sum_{t=1}^{m} \xi_t T(\omega_t)|$ where $\xi_t$ is a sequence of $\mathbb{N}(0,1)$ random variables. With probability at least $1-\delta$ over $\xi=\{\xi_t\}$, we can relate this value to a Gaussian maxima as follows \cite{pisier1999volume}:
\begin{equation*}
\begin{aligned}
\underset{T \in K_M^T(\alpha,R)} {\sup} |\frac{1}{m} \sum_{t=1}^{m} \xi_t T(\omega_t)| &\leq \mathbb{E}_{\xi}[\underset{T \in K_M^T(\alpha,R)} {\sup} |\frac{1}{m} \sum_{t=1}^{m} \xi_t T(\omega_t)|] + \pi \alpha \sqrt{\frac{\log (\frac{1}{\delta})}{2m}} \\
&\leq R\ \mathbb{E}_{\xi}[\underset{T \in T_{\pm}} {\sup} |\frac{1}{m} \sum_{t=1}^{m} \xi_t T(\omega_t)|] + \pi \alpha \sqrt{\frac{\log (\frac{1}{\delta})}{2m}},
\end{aligned}
\end{equation*}
where $T \in T_{\pm}$ is  the set of rank-one sign tensors with $|T_{\pm}| < 2^{Nd}$. Since for each $T$, $\sum_{t=1}^{m} \xi_t T(\omega_t)$ is a Gaussian with mean zero and variance $m$, the expected maxima is bounded by $\sqrt{2m \log(|T_{\pm}|)}$. Gathering all the above information, we end up with the following upper bound with probability larger than $1-\delta$:
\begin{equation}
\underset{T \in K_M^T(\alpha,R)} {\sup} |\frac{1}{m} \sum_{t=1}^{m} \xi_t T(\omega_t)| \leq R \sqrt{\frac{2\log(2)Nd}{m}} + \pi \alpha \sqrt{\frac{\log (\frac{1}{\delta})}{2m}}.
\end{equation}
Choosing $\delta=e^{-\frac{Nd}{2}}$, we get that with probability at least $1-e^{-\frac{Nd}{2}}$
\begin{equation}\label{TC_final_upperbound}
\underset{T \in K_M^T(\alpha,R)} {\sup} |\frac{1}{m} \sum_{t=1}^{m} \xi_t T(\omega_t)| \leq 2(R+\alpha)\sqrt{\frac{Nd}{m}}.
\end{equation}
\subsubsection{Lower bound on left hand side of \eqref{delta_TC_equation}}
In this section, we prove that with high probability, $\frac{1}{m}\sum_{t=1}^{m} T(\omega_t)^2$ does not deviate much from its expectation $\|T\|_{\Pi}^2$. For ease of notation, define $T_S=(T(\omega_1),T(\omega_2),\cdots,T(\omega_m))$ to be the set of chosen samples drawn from $\Pi$ where
$$\|T\|_{\Pi}^2=\frac{1}{m}\mathbb{E}_{S\sim \Pi} \|T_S\|_2^2=\sum_{\omega} \pi_{\omega} T(\omega)^2.$$

We prove that with high probability over the samples,
\begin{equation}\label{T_S_deviation}
\frac{1}{m} \|T_S\|_2^2 \geq \|T\|_{\Pi}^2 - f_{\beta}(m,N,d),
\end{equation}
hold uniformly for all tensors $T \in K_M^T(1,\beta)$. 

\begin{lem}\label{lem_deviation_empirical}
Defining $\Delta(S):=\underset{T \in K_M^T(1,\beta)} {\sup} |\frac{1}{m} \|T_S\|_2^2 - \|T\|_{\Pi}^2|,$ and assuming $N,d >2$ and $m \leq N^d$, there exist a constant $C>20$ such that
$$\mathbb{P}(\Delta(S) > C\beta \sqrt{\frac{Nd}{m}}) \leq e^{\frac{-N}{ln(N)}}.$$
\end{lem}
To prove this lemma, we show that we can bound the $t$-th moment of $\Delta(S)$ as 
\begin{equation}\label{t-moment}
\mathbb{E}_{S \sim \Pi}[\Delta(S)^t] \leq \left( \frac{8 \beta\sqrt{Nd+t\ \text{ln}(m)}}{\sqrt{m}} \right)^t.
\end{equation}
Before stating the proof of this bound, we show how we can use it to prove Lemma \ref{lem_deviation_empirical} by using Markov's inequality.
\begin{equation}
\mathbb{P}(\Delta(S) > C \beta \sqrt{\frac{Nd}{m}}) = \mathbb{P}\left( (\Delta(S))^t > (C \beta \sqrt{\frac{Nd}{m}})^t \right) \leq \frac{\mathbb{E}_{S \sim \Pi}[\Delta(S)^t]}{(C \beta \sqrt{\frac{Nd}{m}})^t}.
\end{equation}
Using \eqref{t-moment} and simplifying we get
$$\mathbb{P}(\Delta(S) > C \beta \sqrt{\frac{Nd}{m}}) \leq \left( \frac{4 \sqrt{Nd + t \text{ln}(m)}}{C \sqrt{Nd}} \right) ^t$$.
Taking $t = \frac{Nd}{\text{ln}(m)}$ and for $C>12$,
$$\mathbb{P}(\Delta(S) > C \beta \sqrt{\frac{Nd}{m}}) \leq  e^{\frac{-Nd}{\text{ln}(m)}} \leq e^{\frac{-N}{\text{ln}(N)}}.$$ \qed

Now we prove \eqref{t-moment} by using some basic techniques of probability in Banach space, including symmetrization and contraction inequality \cite{ledoux2013probability,davenport20141}. Regarding the tensor $T \in \bigotimes_{i=1}^d \mathbb{R}^N$ as a function from $[N] \times [N] \times \cdots \times [N] \rightarrow R$, we define $f_T(\omega_1, \omega_2, \cdots, \omega_d) := T(\omega_1, \omega_2, \cdots, \omega_d)^2$. We are interested in bounding $\Delta(S):=\underset{f_T : T \in K_M^T(1,\beta)} {\sup} |\frac{1}{m} \sum_{i=1}^{m} f_T(\omega_i) - \mathbb{E}(f_T(\omega_i)) |$. A standard symmetrization argument and using contraction principle yields
$$\mathbb{E}_{S \sim \Pi}[\Delta(S)^t] \leq  \mathbb{E}_{S \sim \Pi} \lbrace 2 \mathbb{E}_{\epsilon} [\underset{T \in K_M^T(1,\beta)} {\sup} |\frac{1}{m} \sum_{i=1}^m \epsilon_i T(\omega_i)^2| ] \rbrace^t \leq \mathbb{E}_{S \sim \Pi} \lbrace 4 \mathbb{E}_{\epsilon} [\underset{T \in K_M^T(1,\beta)} {\sup} |\frac{1}{m} \sum_{i=1}^m \epsilon_i T(\omega_i)| ] \rbrace^t.$$
Notice that if $T_{M} \leq \beta$, then $T \in \beta \text{conv}(T_{\pm})$  and therefore
$$\mathbb{E}_{S \sim \Pi}[\Delta(S)^t] \leq \mathbb{E}_{S \sim \Pi} \lbrace 4 \beta \mathbb{E}_{\epsilon} [\underset{T \in T_{\pm}} {\sup} |\frac{1}{m} \sum_{i=1}^m \epsilon_i T(\omega_i)| ] \rbrace^t = \beta^t \mathbb{E}_{S \sim \Pi} \lbrace \mathbb{E}_{\epsilon} [\underset{T \in T_{\pm}} {\sup} |\frac{4}{m} \sum_{i=1}^m \epsilon_i | ] \rbrace^t.$$
To bound the right hand side above notice that for any $\alpha>0$ \cite[Theorem 36]{srebro2004learning}
$$\mathbb{P}_{\epsilon} ( \frac{4}{m} \sum_{i=1}^m \epsilon_i \geq \frac{\alpha}{\sqrt{m}}) = \mathbb{P} \left (\text{Binom}(m,\frac{1}{2}) \geq \frac{m}{2} + \frac{\alpha \sqrt{m}}{8} \right) \leq e^{\frac{-\alpha^2}{16}}.$$
Taking a union bound over $T_{\pm}$, where $|T_{\pm}| \leq 2^{Nd}$ we get
$$\mathbb{P}_{\epsilon} [\underset{T \in T_{\pm}} {\sup} ( |\frac{4}{m} \sum_{i=1}^m \epsilon_i T(\omega_i)|) \geq \frac{\alpha}{\sqrt{m}}] \leq 2^{Nd+1} e^{\frac{-\alpha^2}{16}}.$$
Combining the above result and using Jensen's inequality, when $t>1$
$$\beta^t \mathbb{E}_{S \sim \Pi} \lbrace \mathbb{E}_{\epsilon} [\underset{T \in T_{\pm}} {\sup} |\frac{4}{m} \sum_{i=1}^m \epsilon_i | ] \rbrace^t \leq \beta^t \mathbb{E}_{S \sim \Pi} \lbrace \mathbb{E}_{\epsilon} [\underset{T \in T_{\pm}} {\sup} |\frac{4}{m} \sum_{i=1}^m \epsilon_i | ]^t \rbrace \leq \beta^t \left( (\frac{\alpha}{\sqrt{m}})^t + 4^t 2^{Nd+1} e^{\frac{-\alpha^2}{16}} \right)$$ 
Choosing $\alpha=\sqrt{16 \text{ln} (4 \times 2^{Nd+1}) + 4t\text{ln}(m)}$ and simplifying proves \eqref{t-moment}.\qed

\subsubsection{Gathering the results of \eqref{delta_TC_equation}, \eqref{TC_final_upperbound}, and \eqref{T_S_deviation}} 

Now we combine the upper and lower bounds in the last two sections to prove Theorem \ref{theorem_atomic_TC}. On one hand from \eqref{TC_final_upperbound}, as $\Delta \in K_T(2\alpha,2R)$, we get
$$ \frac{1}{m}\|\Delta_S\|_2^2 \leq 8\sigma (R+\alpha) \sqrt{\frac{Nd}{m}},$$
with probability greater than $1-e^{-\frac{Nd}{2}}$. On the other hand, using Lemma \ref{lem_deviation_empirical} and rescaling, we get
$$\|\Delta\|_{\Pi}^2 \leq \frac{1}{m} \|\Delta_S\|_2^2 + C R \alpha \sqrt{\frac{Nd}{m}}\},$$
with probability greater than $1-e^{\frac{-N}{\text{ln}(N)}}$. The above two inequalities finishes the proof of Theorem \ref{theorem_atomic_TC}.
\begin{rem}\label{remark_proof_maxnorm}
There are only two differences in the proof of theorem \ref{theorem_atomic_TC} and Theorem \ref{theorem_maxqnorm_TC}. First is an extra constant, $c_1 c_2^d$, which shows up in Rademacher complexity of unit max-qnorm ball which changes the constant $C$ in Theorem \ref{theorem_atomic_TC} to $C c_1 c_2^d$ in Theorem \ref{theorem_maxqnorm_TC} and second is the max-qnorm of the error tensor $\Delta=\hat{T}-T^{\sharp}$ (refer to equation \eqref{delta_TC_equation}) which belongs to $K_{\m}^T(2\alpha, 2^{d-1} R)$ instead of $K_{\m}^T(2\alpha, 2R)$.
\end{rem}
\subsection{Proof of Theorem \ref{theorem_atomic_TC_lowerbound}}\label{proof_TC_lowerbound}
\subsubsection*{Packing set construction}
In this section we construct a packing for the set $K_{M}^T(\alpha,R)$.
\begin{lem}\label{packing}
Let $r=\floor{(\frac{R}{\alpha K_G})^2}$ and let $K_{M}^T(\alpha,R)$ be defined as in \eqref{K} and let $\gamma \leq 1$ be such that $\frac{r}{\gamma^2}$ is an integer and suppose $\frac{r}{\gamma^2} \leq N$. Then there exist a set $\chi^T \subset G_{M}^T(\alpha,R)$ with
$$|\chi^T| \geq \exp{\frac{rN}{16\gamma^2}}$$
such that
\begin{enumerate}
\item For $T \in \chi^T$, $|T(\omega)|=\alpha \gamma$ for $\omega \in \{[N] \times [N] \cdots [N]\}$.
\item For any $T^{(i)},T^{(j)} \in \chi^T$, $T^{(i)} \neq T^{(j)}$
$$\|T^{(i)}-T^{(j)}\|_F^2 \geq \frac{\alpha^2 \gamma^2 N^d}{2}.$$
\end{enumerate}
\end{lem}
\noindent
{\bf{Proof:}} This packing is a tensor version of the packing set generated in \cite{davenport20141} with similar properties and our construction is based on the packing set generated there for low-rank matrices with bounded entries. In particular we know that there exist a set $\chi \subset \lbrace M \in \mathbb{R}^{N \times N}: \|M\|_{\infty} \leq \alpha, \text{rank}(M)=r \rbrace$ with $|\chi| \geq \text{exp}{\frac{rN}{16\gamma^2}}$ and for any $M^{(i)},M^{(j)} \in \chi$, $\|M^{(i)}-M^{(j)}\|_F^2 \geq \frac{\alpha^2 \gamma^2 N^2}{2}$ when $i \neq j$. Take any $M^{(k)} \in \chi$. $M^{(k)}$ is a rank-$r$ matrix with $\|M^{(k)}\|_{\infty} \leq \alpha$ and therefore $\|M^{(k)}\|_{\m} \leq \sqrt{r} \alpha$ which means there exist a nuclear decomposition of $M^{(k)}$ with bounded singular vectors $M^{(k)}=\sum_{i} \sigma_i u_i \circ v_i,\ \|u_i\|_{\infty},\|v_i\|_{\infty} \leq 1$, such that $\sum_{i=1} |\sigma_i| \leq K_G \sqrt{r} \alpha$. Define $T^{(k)} = \sum_{i} \sigma_i u_i \circ v_i \circ \vec{\mathbf{1}} \cdots \circ \vec{\mathbf{1}}$ where $\vec{\mathbf{1}} \in \mathbb{R}^N$ is the vector of all ones. Notice that $\|u_i\|_{\infty}, \|v_i\|_{\infty}, \|\vec{\mathbf{1}}\|_{\infty} \leq 1$ and therefore by definition, $\|T^{(k)}\|_{M} \leq K_G \sqrt{r} \alpha \leq R$ by Lemma \ref{max_tensor_ball}. The tensor is basically generated by stacking the matrix $M^{(k)}$ along all the other $d-2$ dimensions and therefore $|T^{(k)}(\omega)|=\alpha \gamma$ for $\omega \in \{[N] \times [N] \cdots [N]\}$, and $\|T^{(k)}\|_{\infty} \leq \alpha$. Hence we build $\chi^T$ from $\chi$ by taking the outer product of the matrices in $\chi$ by the vector $\vec{\mathbf{1}}$ along all the other dimensions. obviously $|\chi^T|=|\chi| \geq \text{exp}{\frac{rN}{16\gamma^2}}$. It just remains to prove $$\|T^{(i)}-T^{(j)}\|_F^2 \geq \frac{\alpha^2 \gamma^2 N^d}{2}$$.

Assuming that $T^{(i)}$ is generated from $M^{(i)}$ and $T^{(j)}$ is generated from $M^{(j)}$, since $T^{(i)}(i_1,i_2, \cdots,i_d)=M^{(i)}(i_1,i_2)$, $\|T^{(i)}-T^{(j)}\|_F^2 = \sum_{i_1=1}^{N} \sum_{i_2=1}^{N} \cdots \sum_{i_d=1}^{N} (T^{(i)}(i_1,i_2, \cdots,i_d) - T^{(j)}(i_1,i_2, \cdots,i_d))^2= \\ N^{d-2} \sum_{i_1=1}^{N} \sum_{i_2=1}^{N} (M^{(i)}(i_1,i_2)-M^{(j)}(i_1,i_2))^2 = N^{d-2} \|M^{(i)}-M^{(j)}\|_F^2 \geq \frac{\alpha^2 \gamma^2 N^d}{2}$ which concludes proof of the lemma.
\subsubsection*{Proof of Theorem \ref{theorem_atomic_TC_lowerbound}}
Now we use the construction in Lemma \ref{packing} to obtain a $\delta$-packing set, $\chi^T$ of $K_{M}^T$, with $\delta=\alpha \gamma \sqrt{\frac{N^d}{2}}$. For the lower bound we assume that the sampling distribution satisfies 
\begin{equation}\label{upperbound_on_pi}
\max_{\omega} \pi_{\omega} \leq \frac{L}{N^d}.
\end{equation}
The proof is based on the proof in \cite[Section 6.2]{cai2016matrix} which we will rewrite the main parts and refer to \cite{cai2016matrix} for more details. The proof is based on two main arguments. First is a lower bound on the $\|\cdot\|_F$-risk in terms of the error in a multi-way hypothesis testing problem
$$\underset{\hat{T}}{\inf} \underset{T \in G_{M}^T(\alpha,R)}{\sup} \mathbb{E}\|\hat{T}-T\|_F^2 \geq \frac{\delta^2}{4} \underset{\tilde{T}}{\text{min}} \mathbb{P} (\tilde{T} \neq T^{\sharp}),$$
where $T^{\sharp}$ is uniformly distributed over the pacing set $\chi^T$. Second argument is a variant of the Fano's inequality which conditioned on the observations $S=\{\omega_1, \cdots, \omega_m\}$, gives the lower bound
\begin{equation}\label{Fano_upper_bound}
\mathbb{P}(\tilde{T} \neq T^{\sharp} | S) \geq 1-\frac{(\binom{|\chi^T|}{2})^{-1} \sum_{k\neq j} K(T^k || T^j)+\log(2)}{\log|\chi^T|},
\end{equation}
where $K(T^k || T^j)$ is the Kullback-Leibler divergence between distributions $(Y_S|T^k)$ and $(Y_S|T^j)$. For our observation model with i.i.d. Gaussian noise, we have
$$K(T^k || T^j) = \frac{1}{2\sigma^2}\sum_{t=1}^{m} (T^k(\omega_t) - T^j(\omega_t))^2,$$
and therefore,
$$\mathbb{E}_S [K(T^k || T^j)] = \frac{m}{2\sigma^2} \|T^k - T^j\|_{\Pi}^2.$$

From the first property of the packing set generated in Lemma \ref{packing}, $\|T^k-T^j\|_F^2 \leq 4 \gamma^2 N^d$. This combined \eqref{upperbound_on_pi} and \eqref{Fano_upper_bound} yields
$$\mathbb{P}(\tilde{T} \neq T^{\sharp}) \geq 1-\frac{\frac{32L\gamma^4\alpha^2m}{\sigma^2}+12\gamma^2}{rN}\geq 1-\frac{32L\gamma^4\alpha^2m}{\sigma^2rN}-\frac{12\gamma^2}{rN}\geq \frac{3}{4}-\frac{32L\gamma^4\alpha^2m}{\sigma^2rN},$$
provided $rN>48$. Now if $\gamma^4 \leq \frac{\sigma^2rN}{128L\alpha^2m}$,
$$\underset{\hat{T}}{\inf} \underset{T \in K_{M}^T(\alpha,R)}{\sup} \frac{1}{N^d} \mathbb{E}\|\hat{T}-T\|_F^2 \geq \frac{\alpha^2 \gamma^2}{16}.$$

Therefore, if $\frac{\sigma^2rN}{128L\alpha^2m} \geq 1$ choosing $\gamma=1$ finishes the proof. Otherwise choosing $\gamma^2 = \frac{\sigma}{8\sqrt{2}\alpha}\sqrt{\frac{rN}{Lm}}$ results in
$$\underset{\hat{T}}{\inf} \underset{T \in K_{M}^T(\alpha,R)}{\sup} \frac{1}{N^d}  \mathbb{E}\|\hat{T}-T\|_F^2 \geq \frac{\sigma\alpha}{128\sqrt{2}}\sqrt{\frac{rN}{Lm}} \geq \frac{\sigma R}{128\sqrt{2}K_G}\sqrt{\frac{N}{Lm}}.$$

\section{Future directions and open problems}
In this work we considered max-qnorm constrained least squares for tensor completion and showed that theoretically the number of required measurements needed is linear in the maximum size of the tensor. To the best of our knowledge, this is the first work that reduces the required number of measurements from $N^{\frac{d}{2}}$ to $N$. However, there are a lot of open problems and complications that need to be answered. Following, we list a few of these problems.
\begin{itemize}
\item The difference between the upper bound of nuclear-norm and max-qnorm of a bounded low-rank tensor is significant and it is also a main reason for the theoretical superiority of max-qnorm over nuclear-norm. In our proof, one of the main theoretical steps for bounding the least square estimation error, constrained with an arbitrary norm, is bounding the Rademacher complexity of unit-norm tensors and finding a tight bound for the norm of low-rank tensors. In case of max-qnorm, we are able to achieve upper bound of $r^{\sqrt{d^2-d}} \alpha$ and Rademacher complexity of $O(\sqrt{\frac{dN}{m}})$. A careful calculation of these quantities for nuclear-norm still needs to be done. However, a generalizations of current results gives an upper bound of $O(\sqrt{r^{d-1} N^d} \alpha)$ for the nuclear-norm of rank-$r$ tensors. Considering the tensor $\vec{\mathbf{1}} \circ \cdots \circ \vec{\mathbf{1}} $, we can see that this bound is tight. We leave the exact answer to this question to future work.
\item We know that the dependency of the upper bound of the low-rank max-qnorm tensor found in Theorem \ref{theorem_atomicnorm_bound} is optimal in $N$. However, we believe the dependency on $r$ can be improved. We saw in Section \ref{experiments and algorithms} that this is definitely the case for some specific class of tensors.
\item Other than the open problems concerning algorithms for calculating max-qnorm of tensors and projection onto max-qnorm balls, an interesting question is analyzing exact tensor recovery using max-qnorm. Most of the evidence point to this being NP-hard including \cite{hillar2013most} which proves that a lot of similar tensor problems is NP-hard and the connection between noisy tensor completion and the 3-SAT problem \cite{barak2015noisy} which proves that if an exact tensor completion is doable in polynomial time, the conjecture in \cite{daniely2013more} will be disapproved. However, an exact study of whether or not it is NP-hard or availability of polynomial time estimates needs to be done.
\item The preliminary results of Algorithm \ref{algorithm_TC} show significant improvements over previous algorithms. This highlights the need for a more sophisticated (and provable) algorithm which utilizes the max-qnorm for tensor completion. As a first step, in the matrix case, the max-qnorm can be reformulated as a semidefinite programming problem, and together with \cite{burer2006computational}, this proves once we solve the problem using its factors, any local minima is a global minima and hence proves the correctness of the algorithm. However, this is not the case in tensors and in our experiments we saw that the results are sensitive to the size of low-rank factors. Analyzing this behavior is an interesting future direction.
\end{itemize} 
\bibliography{sparse}

\begin{thebibliography}{10}

\bibitem{acar2005modeling}
Evrim Acar, Seyit~A {\c{C}}amtepe, Mukkai~S Krishnamoorthy, and B{\"u}lent
  Yener.
\newblock Modeling and multiway analysis of chatroom tensors.
\newblock In {\em International Conference on Intelligence and Security
  Informatics}, pages 256--268. Springer, 2005.

\bibitem{barak2015noisy}
Boaz Barak and Ankur Moitra.
\newblock Noisy tensor completion via the sum-of-squares hierarchy.
\newblock {\em arXiv preprint arXiv:1501.06521}, 2015.

\bibitem{bartlett2002rademacher}
Peter~L Bartlett and Shahar Mendelson.
\newblock Rademacher and gaussian complexities: Risk bounds and structural
  results.
\newblock {\em Journal of Machine Learning Research}, 3(Nov):463--482, 2002.

\bibitem{bazerque2013rank}
Juan~Andr{\'e}s Bazerque, Gonzalo Mateos, and Georgios Giannakis.
\newblock Rank regularization and bayesian inference for tensor completion and
  extrapolation.
\newblock {\em Signal Processing, IEEE Transactions on}, 61(22):5689--5703,
  2013.

\bibitem{bhojanapalli2014universal}
Srinadh Bhojanapalli and Prateek Jain.
\newblock Universal matrix completion.
\newblock {\em arXiv preprint arXiv:1402.2324}, 2014.

\bibitem{blei1979multidimensional}
Ron~C Blei.
\newblock Multidimensional extensions of the grothendieck inequality and
  applications.
\newblock {\em Arkiv f{\"o}r Matematik}, 17(1):51--68, 1979.

\bibitem{burer2006computational}
Samuel Burer and Changhui Choi.
\newblock Computational enhancements in low-rank semidefinite programming.
\newblock {\em Optimisation Methods and Software}, 21(3):493--512, 2006.

\bibitem{cai2010singular}
Jian-Feng Cai, Emmanuel~J Cand{\`e}s, and Zuowei Shen.
\newblock A singular value thresholding algorithm for matrix completion.
\newblock {\em SIAM Journal on Optimization}, 20(4):1956--1982, 2010.

\bibitem{cai2016matrix}
T~Tony Cai, Wen-Xin Zhou, et~al.
\newblock Matrix completion via max-norm constrained optimization.
\newblock {\em Electronic Journal of Statistics}, 10(1):1493--1525, 2016.

\bibitem{cai2013max}
Tony Cai and Wen-Xin Zhou.
\newblock A max-norm constrained minimization approach to 1-bit matrix
  completion.
\newblock {\em Journal of Machine Learning Research}, 14(1):3619--3647, 2013.

\bibitem{candes2010matrix}
Emmanuel~J Candes and Yaniv Plan.
\newblock Matrix completion with noise.
\newblock {\em Proceedings of the IEEE}, 98(6):925--936, 2010.

\bibitem{candes2009exact}
Emmanuel~J Cand{\`e}s and Benjamin Recht.
\newblock Exact matrix completion via convex optimization.
\newblock {\em Foundations of Computational mathematics}, 9(6):717--772, 2009.

\bibitem{candes2010power}
Emmanuel~J Cand{\`e}s and Terence Tao.
\newblock The power of convex relaxation: Near-optimal matrix completion.
\newblock {\em IEEE Transactions on Information Theory}, 56(5):2053--2080,
  2010.

\bibitem{carroll1970analysis}
J~Douglas Carroll and Jih-Jie Chang.
\newblock Analysis of individual differences in multidimensional scaling via an
  n-way generalization of ``eckart-young'' decomposition.
\newblock {\em Psychometrika}, 35(3):283--319, 1970.

\bibitem{chandrasekaran2012convex}
Venkat Chandrasekaran, Benjamin Recht, Pablo~A Parrilo, and Alan~S Willsky.
\newblock The convex geometry of linear inverse problems.
\newblock {\em Foundations of Computational mathematics}, 12(6):805--849, 2012.

\bibitem{da2015optimization}
Curt Da~Silva and Felix~J Herrmann.
\newblock Optimization on the hierarchical tucker manifold--applications to
  tensor completion.
\newblock {\em Linear Algebra and its Applications}, 481:131--173, 2015.

\bibitem{daniely2013more}
Amit Daniely, Nati Linial, and Shai Shalev-Shwartz.
\newblock More data speeds up training time in learning halfspaces over sparse
  vectors.
\newblock In {\em Advances in Neural Information Processing Systems}, pages
  145--153, 2013.

\bibitem{davenport20141}
Mark~A Davenport, Yaniv Plan, Ewout van~den Berg, and Mary Wootters.
\newblock 1-bit matrix completion.
\newblock {\em Information and Inference}, 3(3):189--223, 2014.

\bibitem{davenport2016overview}
Mark~A Davenport and Justin Romberg.
\newblock An overview of low-rank matrix recovery from incomplete observations.
\newblock {\em IEEE Journal of Selected Topics in Signal Processing},
  10(4):608--622, 2016.

\bibitem{derksen2013nuclear}
Harm Derksen.
\newblock On the nuclear norm and the singular value decomposition of tensors.
\newblock {\em Foundations of Computational Mathematics}, pages 1--33, 2013.

\bibitem{fang2015max}
Ethan~X Fang, Han Liu, Kim-Chuan Toh, and Wen-Xin Zhou.
\newblock Max-norm optimization for robust matrix recovery.
\newblock {\em Mathematical Programming}, pages 1--31, 2015.

\bibitem{fazel2002matrix}
Maryam Fazel.
\newblock {\em Matrix rank minimization with applications}.
\newblock PhD thesis, PhD thesis, Stanford University, 2002.

\bibitem{foygel2011concentration}
Rina Foygel and Nathan Srebro.
\newblock Concentration-based guarantees for low-rank matrix reconstruction.
\newblock In {\em COLT}, pages 315--340, 2011.

\bibitem{foygel2012matrix}
Rina Foygel, Nathan Srebro, and Ruslan~R Salakhutdinov.
\newblock Matrix reconstruction with the local max norm.
\newblock In {\em Advances in Neural Information Processing Systems}, pages
  935--943, 2012.

\bibitem{friedland1982variation}
Shmuel Friedland.
\newblock Variation of tensor powers and spectrat.
\newblock {\em Linear and Multilinear algebra}, 12(2):81--98, 1982.

\bibitem{gandy2011tensor}
Silvia Gandy, Benjamin Recht, and Isao Yamada.
\newblock Tensor completion and low-n-rank tensor recovery via convex
  optimization.
\newblock {\em Inverse Problems}, 27(2):025010, 2011.

\bibitem{grasedyck2013literature}
Lars Grasedyck, Daniel Kressner, and Christine Tobler.
\newblock A literature survey of low-rank tensor approximation techniques.
\newblock {\em GAMM-Mitteilungen}, 36(1):53--78, 2013.

\bibitem{grothendieck1955produits}
Alexandre Grothendieck.
\newblock Produits tensoriels topologiques et espaces nucl{\'e}aires.
\newblock {\em S{\'e}minaire Bourbaki}, 2:193--200, 1955.

\bibitem{harshman1970foundations}
Richard~A Harshman.
\newblock {\em Foundations of the parafac procedure: models and conditions for
  an ``explanatory" multimodal factor analysis}.
\newblock University of California at Los Angeles Los Angeles, CA, 1970.

\bibitem{hillar2013most}
Christopher~J Hillar and Lek-Heng Lim.
\newblock Most tensor problems are np-hard.
\newblock {\em Journal of the ACM (JACM)}, 60(6):45, 2013.

\bibitem{hu2015relations}
Shenglong Hu.
\newblock Relations of the nuclear norm of a tensor and its matrix flattenings.
\newblock {\em Linear Algebra and its Applications}, 478:188--199, 2015.

\bibitem{jain2013low}
Prateek Jain, Praneeth Netrapalli, and Sujay Sanghavi.
\newblock Low-rank matrix completion using alternating minimization.
\newblock In {\em Proceedings of the forty-fifth annual ACM symposium on Theory
  of computing}, pages 665--674. ACM, 2013.

\bibitem{jain2014provable}
Prateek Jain and Sewoong Oh.
\newblock Provable tensor factorization with missing data.
\newblock In {\em Advances in Neural Information Processing Systems}, pages
  1431--1439, 2014.

\bibitem{john2014extremum}
Fritz John.
\newblock Extremum problems with inequalities as subsidiary conditions.
\newblock In {\em Traces and emergence of nonlinear programming}, pages
  197--215. Springer, 2014.

\bibitem{keshavan2010matrix}
Raghunandan~H Keshavan, Andrea Montanari, and Sewoong Oh.
\newblock Matrix completion from noisy entries.
\newblock {\em Journal of Machine Learning Research}, 11(Jul):2057--2078, 2010.

\bibitem{keshavan2009matrix}
Raghunandan~H Keshavan, Sewoong Oh, and Andrea Montanari.
\newblock Matrix completion from a few entries.
\newblock In {\em 2009 IEEE International Symposium on Information Theory},
  pages 324--328. IEEE, 2009.

\bibitem{kolda2009tensor}
Tamara~G Kolda and Brett~W Bader.
\newblock Tensor decompositions and applications.
\newblock {\em SIAM review}, 51(3):455--500, 2009.

\bibitem{kolda2006multilinear}
Tamara~Gibson Kolda.
\newblock {\em Multilinear operators for higher-order decompositions}.
\newblock United States. Department of Energy, 2006.

\bibitem{kreimer2013tensor}
Nadia Kreimer, Aaron Stanton, and Mauricio~D Sacchi.
\newblock Tensor completion based on nuclear norm minimization for 5d seismic
  data reconstruction.
\newblock {\em Geophysics}, 78(6):V273--V284, 2013.

\bibitem{krishnamurthy2013low}
Akshay Krishnamurthy and Aarti Singh.
\newblock Low-rank matrix and tensor completion via adaptive sampling.
\newblock In {\em Advances in Neural Information Processing Systems}, pages
  836--844, 2013.

\bibitem{kushner2012stochastic}
Harold~Joseph Kushner and Dean~S Clark.
\newblock {\em Stochastic approximation methods for constrained and
  unconstrained systems}, volume~26.
\newblock Springer Science \& Business Media, 2012.

\bibitem{ledoux2013probability}
Michel Ledoux and Michel Talagrand.
\newblock {\em Probability in Banach Spaces: isoperimetry and processes}.
\newblock Springer Science \& Business Media, 2013.

\bibitem{lee2010practical}
Jason~D Lee, Ben Recht, Nathan Srebro, Joel Tropp, and Ruslan~R Salakhutdinov.
\newblock Practical large-scale optimization for max-norm regularization.
\newblock In {\em Advances in Neural Information Processing Systems}, pages
  1297--1305, 2010.

\bibitem{li2010tensor}
Nan Li and Baoxin Li.
\newblock Tensor completion for on-board compression of hyperspectral images.
\newblock In {\em 2010 IEEE International Conference on Image Processing},
  pages 517--520. IEEE, 2010.

\bibitem{lim2010multiarray}
Lek-Heng Lim and Pierre Comon.
\newblock Multiarray signal processing: Tensor decomposition meets compressed
  sensing.
\newblock {\em Comptes Rendus Mecanique}, 338(6):311--320, 2010.

\bibitem{linial2007complexity}
Nati Linial, Shahar Mendelson, Gideon Schechtman, and Adi Shraibman.
\newblock Complexity measures of sign matrices.
\newblock {\em Combinatorica}, 27(4):439--463, 2007.

\bibitem{liu2013tensor}
Ji~Liu, Przemyslaw Musialski, Peter Wonka, and Jieping Ye.
\newblock Tensor completion for estimating missing values in visual data.
\newblock {\em IEEE Transactions on Pattern Analysis and Machine Intelligence},
  35(1):208--220, 2013.

\bibitem{ma2011fixed}
Shiqian Ma, Donald Goldfarb, and Lifeng Chen.
\newblock Fixed point and bregman iterative methods for matrix rank
  minimization.
\newblock {\em Mathematical Programming}, 128(1-2):321--353, 2011.

\bibitem{mocks1988topographic}
J~Mocks.
\newblock Topographic components model for event-related potentials and some
  biophysical considerations.
\newblock {\em IEEE transactions on biomedical engineering}, 6(35):482--484,
  1988.

\bibitem{mu2014square}
Cun Mu, Bo~Huang, John Wright, and Donald Goldfarb.
\newblock Square deal: Lower bounds and improved relaxations for tensor
  recovery.
\newblock In {\em Proceedings of the 31st International Conference on Machine
  Learning (ICML-14)}, pages 73--81, 2014.

\bibitem{negahban2012restricted}
Sahand Negahban and Martin~J Wainwright.
\newblock Restricted strong convexity and weighted matrix completion: Optimal
  bounds with noise.
\newblock {\em Journal of Machine Learning Research}, 13(May):1665--1697, 2012.

\bibitem{nion2010tensor}
Dimitri Nion and Nicholas~D Sidiropoulos.
\newblock Tensor algebra and multidimensional harmonic retrieval in signal
  processing for mimo radar.
\newblock {\em IEEE Transactions on Signal Processing}, 58(11):5693--5705,
  2010.

\bibitem{pisier1999volume}
Gilles Pisier.
\newblock {\em The volume of convex bodies and Banach space geometry},
  volume~94.
\newblock Cambridge University Press, 1999.

\bibitem{rashtchian2016bounded}
Cyrus Rashtchian.
\newblock Bounded matrix rigidity and john's theorem.
\newblock In {\em Electronic Colloquium on Computational Complexity (ECCC)},
  volume~23, page~93, 2016.

\bibitem{recht2011simpler}
Benjamin Recht.
\newblock A simpler approach to matrix completion.
\newblock {\em Journal of Machine Learning Research}, 12(Dec):3413--3430, 2011.

\bibitem{rockafellar2015convex}
Ralph~Tyrell Rockafellar.
\newblock {\em Convex analysis}.
\newblock Princeton university press, 2015.

\bibitem{schatten1985theory}
Robert Schatten.
\newblock {\em A theory of cross-spaces}.
\newblock Number~26. Princeton University Press, 1985.

\bibitem{schmidt2009optimizing}
Mark~W Schmidt, Ewout Van Den~Berg, Michael~P Friedlander, and Kevin~P Murphy.
\newblock Optimizing costly functions with simple constraints: A limited-memory
  projected quasi-newton algorithm.
\newblock In {\em AISTATS}, volume~5, page 2009, 2009.

\bibitem{shashua2001linear}
Amnon Shashua and Anat Levin.
\newblock Linear image coding for regression and classification using the
  tensor-rank principle.
\newblock In {\em Computer Vision and Pattern Recognition, 2001. CVPR 2001.
  Proceedings of the 2001 IEEE Computer Society Conference on}, volume~1, pages
  I--42. IEEE, 2001.

\bibitem{shen2014online}
Jie Shen, Huan Xu, and Ping Li.
\newblock Online optimization for max-norm regularization.
\newblock In {\em Advances in Neural Information Processing Systems}, pages
  1718--1726, 2014.

\bibitem{signoretto2010nuclear}
Marco Signoretto, Lieven De~Lathauwer, and Johan~AK Suykens.
\newblock Nuclear norms for tensors and their use for convex multilinear
  estimation.
\newblock {\em Submitted to Linear Algebra and Its Applications}, 43, 2010.

\bibitem{srebro2004learning}
Nathan Srebro.
\newblock {\em Learning with matrix factorizations}.
\newblock PhD thesis, Citeseer, 2004.

\bibitem{srebro2005maximum}
Nathan Srebro, Jason Rennie, and Tommi~S Jaakkola.
\newblock Maximum-margin matrix factorization.
\newblock In {\em Advances in neural information processing systems}, pages
  1329--1336, 2005.

\bibitem{srebro2005rank}
Nathan Srebro and Adi Shraibman.
\newblock Rank, trace-norm and max-norm.
\newblock In {\em International Conference on Computational Learning Theory},
  pages 545--560. Springer, 2005.

\bibitem{tomioka2010estimation}
Ryota Tomioka, Kohei Hayashi, and Hisashi Kashima.
\newblock Estimation of low-rank tensors via convex optimization.
\newblock {\em arXiv preprint arXiv:1010.0789}, 2010.

\bibitem{tonge1978neumann}
Andrew Tonge.
\newblock The von neumann inequality for polynomials in several hilbert-schmidt
  operators.
\newblock {\em Journal of the London Mathematical Society}, 2(3):519--526,
  1978.

\bibitem{tucker1966some}
Ledyard~R Tucker.
\newblock Some mathematical notes on three-mode factor analysis.
\newblock {\em Psychometrika}, 31(3):279--311, 1966.

\bibitem{van2013fast}
Tristan van Leeuwen and Felix~J Herrmann.
\newblock Fast waveform inversion without source-encoding.
\newblock {\em Geophysical Prospecting}, 61(s1):10--19, 2013.

\bibitem{xu2012alternating}
Yangyang Xu, Wotao Yin, Zaiwen Wen, and Yin Zhang.
\newblock An alternating direction algorithm for matrix completion with
  nonnegative factors.
\newblock {\em Frontiers of Mathematics in China}, 7(2):365--384, 2012.

\bibitem{yuan2015tensor}
Ming Yuan and Cun-Hui Zhang.
\newblock On tensor completion via nuclear norm minimization.
\newblock {\em Foundations of Computational Mathematics}, pages 1--38, 2015.

\bibitem{yuan2016tensor}
Ming Yuan and Cun-Hui Zhang.
\newblock On tensor completion via nuclear norm minimization.
\newblock {\em Foundations of Computational Mathematics}, 16(4):1031--1068,
  2016.

\bibitem{zhang2015exact}
Zemin Zhang and Shuchin Aeron.
\newblock Exact tensor completion using t-svd.
\newblock {\em arXiv preprint arXiv:1502.04689}, 2015.

\end{thebibliography}
\bibliographystyle{plain}
\begin{appendices}
\section{Rademacher Complexity}\label{rademacher_complexity}
A technical tool that we use in the proof of our main results involves data-dependent estimates of the Rademacher
and Gaussian complexities of a function class. We refer to \cite{bartlett2002rademacher,srebro2005rank} for a detailed introduction of these concepts.
\begin{dfn}\cite{cai2013max}
Let $\mathbb{P}$ be a probability distribution on a set $\chi$ and assume the set $S:=\lbrace X_1,\cdots, X_m\rbrace$ is $m$ independent samples drawn from $\chi$ according to $\mathbb{P}$. The empirical Rademacher complexity of a class of Functions $\mathbb{F}$ defined from $\chi$ to $\mathbb{R}$ is defined as
$$\hat{R}_S(\mathbb{F})=\frac{2}{|S|}\mathbb{E}_{\epsilon}[\underset{f \in \mathbb{F}}{sup}|\sum_{i=1}^{i=m}\epsilon_if(X_i)|],$$
where $\epsilon_i$ is a Rademacher random variable. Moreover, the Rademacher complexity with respect to the distribution $\mathbb{P}$ over a sample $S$ of $|S|$ points drawn independently according to $\mathbb{P}$ is defined as the expectation of the empirical Rademacher complexity defined as:
$$R_{|S|}(\mathbb{F})=\mathbb{E}_{S\sim \mathbb{P}}[\hat{R}_S(\mathbb{F})]$$
\end{dfn} 
Two important properties that will be used in the following lemma is: First, if $\mathbb{F} \subset \mathbb{G}$, then$\hat{R}_S(\mathbb{F}) \leq \hat{R}_S(\mathbb{G})$ and second is $\hat{R}_S(\mathbb{F})=\hat{R}_S(\mathbb{\text{conv}(F)})$.
\begin{lem}\label{Rademacher_atomicnorm}
$\underset{S:|S|=m}{sup} \hat{R}_S(\mathbb{B}_{M}(1)) < 6 \sqrt{\frac{dN}{m}}$ 
\end{lem}
\noindent
\textbf{Proof}: By definition, $\mathbb{B}_{M}(1)=\text{conv}(T_{\pm})$, and $T_{\pm}$ is a finite class with $|T_{\pm}|<2^{dN}$. Therefore, $\hat{R}_S(T_{\pm}) < \sqrt{7\frac{2dN+log|S|}{|S|}}$ \cite{srebro2004learning} which concludes the proof.
\begin{lem}\label{Rademacher_maxnorm}
$\underset{S:|S|=m}{sup} \hat{R}_S(\mathbb{B}_{\m}^T(1)) < 6 c_1 c_2^d \sqrt{\frac{dN}{m}}$ 
\end{lem}
\noindent
\textbf{Proof}: By Lemma \ref{max_tensor_ball}, $\mathbb{B}_{\m}^T(1)$ $\subset$ $c_1 c_2^d$ conv($T_{\pm}$) and we have $\hat{R}_S(T_{\pm}) < \sqrt{7\frac{2dN+log|S|}{|S|}}$. Taking the convex hull of this class and using $|S|=m\leq N^d$ and scaling by $c_1 c_2^d$ we get $\hat{R}_S(\mathbb{B}_{\m}^T(1)) \leq 6 c_1 c_2^d \sqrt{\frac{dN}{m}}$.
\end{appendices}
\end{document}